\documentclass[final,5p,times,twocolumn,authoryear,nopreprintline]{elsarticle}

\usepackage{amssymb}

\usepackage[table, dvipsnames]{xcolor}
\usepackage{bm}
\usepackage{soul}
\usepackage{amsmath}
\usepackage{amsfonts}
\usepackage{amssymb}
\usepackage{algorithm}
\usepackage[noend]{algpseudocode}
\usepackage{tabularx}
\usepackage{caption}
\usepackage{multirow}
\usepackage{dblfloatfix}
\usepackage{hyperref}
\usepackage{pgfplots}
\pgfplotsset{compat=1.17}

\usepackage{enumitem}
\usepackage{times}
\usepackage{epsfig}
\usepackage{graphicx}
\usepackage{amsmath, bm}
\usepackage{amssymb}
\usepackage{booktabs}
\usepackage{algorithm,algorithmicx,algpseudocode}
\usepackage{subfig}
\usepackage{multirow}
\usepackage{xspace}
\usepackage{url}
\usepackage{graphicx}
\usepackage{adjustbox}

\journal{Elsevier}

\begin{document}

\begin{frontmatter}




\definecolor{lightblue}{RGB}{135,206,250}
\title{UB-FineNet: \textcolor{lightblue}{U}rban \textcolor{lightblue}{B}uilding \textcolor{lightblue}{Fine}-grained Classification \textcolor{lightblue}{Net}work for Open-access Satellite Images}

\author[polyu]{Zhiyi He}

\author[polyu,polyusri,polyuscri]{Wei Yao\corref{cor1}}
\ead{wei.hn.yao@polyu.edu.hk}

\author[polyu,polyusri]{Jie Shao}

\author[polyu]{Puzuo Wang}

\address[polyu]{Department of Land Surveying and Geo-Informatics, The Hong Kong Polytechnic University, Hong Kong}
\address[polyusri]{The Hong Kong Polytechnic University Shenzhen Research Institute, Shenzhen, China}
\address[polyuscri]{Otto Poon Charitable Foundation Smart Cities Research Institute, The Hong Kong Polytechnic University, Hong Kong}

\cortext[cor1]{Corresponding author.}

\begin{abstract}

Fine classification of city-scale buildings from satellite remote sensing imagery is a crucial research area with significant implications for urban planning, infrastructure development, and population distribution analysis. However, the task faces big challenges due to low-resolution overhead images acquired from high altitude space-borne platforms and the long-tail sample distribution of fine-grained urban building categories, leading to severe class imbalance problem. To address these issues, we propose an deep network approach to fine-grained classification of urban buildings using open-access satellite images. A Denoising Diffusion Probabilistic Model (DDPM) based super-resolution method is first introduced to enhance the spatial resolution of satellite images, which benefits from domain-adaptive knowledge distillation. Then, a new fine-grained classification network with Category Information Balancing Module (CIBM) and Contrastive Supervision (CS) technique is proposed to mitigate the problem of class imbalance and improve the classification robustness and accuracy. Experiments on Hong Kong data set with 11 fine building types revealed promising classification results with a mean Top-1 accuracy of 60.45\%,  which is on par with
street-view image based approaches. Extensive ablation study shows that CIBM and CS improve Top-1 accuracy by 2.6\% and 3.5\% compared to the baseline method, respectively. And the both modules can be easily inserted into other classification networks and similar enhancements have been achieved. Our research contributes to the field of urban analysis by providing a practical solution for fine classification of buildings in challenging mega city scenarios solely using open-access satellite images. The proposed method can serve as a valuable tool for urban planners, aiding in the understanding of economic, industrial, and population distribution within cities and regions,
ultimately facilitating informed decision-making processes in urban development and infrastructure planning.

\end{abstract}


\end{frontmatter}



\def\eg{{\em e.g.,}\xspace}
\def\ie{{\em i.e.,}\xspace}
\def\vs{{\em v.s.}\xspace}
\def\cf{{\em c.f.,}\xspace}

\newcommand{\red}[1]{{\textcolor{red}{#1}}}

\newcommand{\figleft}{{\em (Left)}}
\newcommand{\figcenter}{{\em (Center)}}
\newcommand{\figright}{{\em (Right)}}
\newcommand{\figtop}{{\em (Top)}}
\newcommand{\figbottom}{{\em (Bottom)}}
\newcommand{\captiona}{{\em (a)}}
\newcommand{\captionb}{{\em (b)}}
\newcommand{\captionc}{{\em (c)}}
\newcommand{\captiond}{{\em (d)}}

\newcommand{\newterm}[1]{{\bf #1}}

\def\figref#1{Figure~\ref{#1}}
\def\Figref#1{Figure~\ref{#1}}
\def\twofigref#1#2{figures \ref{#1} and \ref{#2}}
\def\quadfigref#1#2#3#4{figures \ref{#1}, \ref{#2}, \ref{#3} and \ref{#4}}
\def\tabref#1{Table~\ref{#1}}
\def\Tabref#1{Table~\ref{#1}}
\def\secref#1{section~\ref{#1}}
\def\Secref#1{Section~\ref{#1}}
\def\twosecrefs#1#2{sections \ref{#1} and \ref{#2}}
\def\secrefs#1#2#3{sections \ref{#1}, \ref{#2} and \ref{#3}}
\def\eqref#1{(\ref{#1})}
\def\Eqref#1{Equation~\ref{#1}}
\def\plaineqref#1{\ref{#1}}
\def\chapref#1{chapter~\ref{#1}}
\def\Chapref#1{Chapter~\ref{#1}}
\def\rangechapref#1#2{chapters\ref{#1}--\ref{#2}}
\def\algref#1{algorithm~\ref{#1}}
\def\Algref#1{Algorithm~\ref{#1}}
\def\twoalgref#1#2{Algorithms \ref{#1} and \ref{#2}}
\def\Twoalgref#1#2{Algorithms \ref{#1} and \ref{#2}}
\def\partref#1{part~\ref{#1}}
\def\Partref#1{Part~\ref{#1}}
\def\twopartref#1#2{parts \ref{#1} and \ref{#2}}

\def\ceil#1{\lceil #1 \rceil}
\def\floor#1{\lfloor #1 \rfloor}
\def\1{\bm{1}}
\newcommand{\train}{\mathcal{D}}
\newcommand{\valid}{\mathcal{D_{\mathrm{valid}}}}
\newcommand{\test}{\mathcal{D_{\mathrm{test}}}}

\def\eps{{\epsilon}}

\def\reta{{\textnormal{$\eta$}}}
\def\ra{{\textnormal{a}}}
\def\rb{{\textnormal{b}}}
\def\rc{{\textnormal{c}}}
\def\rd{{\textnormal{d}}}
\def\re{{\textnormal{e}}}
\def\rf{{\textnormal{f}}}
\def\rg{{\textnormal{g}}}
\def\rh{{\textnormal{h}}}
\def\ri{{\textnormal{i}}}
\def\rj{{\textnormal{j}}}
\def\rk{{\textnormal{k}}}
\def\rl{{\textnormal{l}}}
\def\rn{{\textnormal{n}}}
\def\ro{{\textnormal{o}}}
\def\rp{{\textnormal{p}}}
\def\rq{{\textnormal{q}}}
\def\rr{{\textnormal{r}}}
\def\rs{{\textnormal{s}}}
\def\rt{{\textnormal{t}}}
\def\ru{{\textnormal{u}}}
\def\rv{{\textnormal{v}}}
\def\rw{{\textnormal{w}}}
\def\rx{{\textnormal{x}}}
\def\ry{{\textnormal{y}}}
\def\rz{{\textnormal{z}}}

\def\rvepsilon{{\mathbf{\epsilon}}}
\def\rvtheta{{\mathbf{\theta}}}
\def\rva{{\mathbf{a}}}
\def\rvb{{\mathbf{b}}}
\def\rvc{{\mathbf{c}}}
\def\rvd{{\mathbf{d}}}
\def\rve{{\mathbf{e}}}
\def\rvf{{\mathbf{f}}}
\def\rvg{{\mathbf{g}}}
\def\rvh{{\mathbf{h}}}
\def\rvu{{\mathbf{i}}}
\def\rvj{{\mathbf{j}}}
\def\rvk{{\mathbf{k}}}
\def\rvl{{\mathbf{l}}}
\def\rvm{{\mathbf{m}}}
\def\rvn{{\mathbf{n}}}
\def\rvo{{\mathbf{o}}}
\def\rvp{{\mathbf{p}}}
\def\rvq{{\mathbf{q}}}
\def\rvr{{\mathbf{r}}}
\def\rvs{{\mathbf{s}}}
\def\rvt{{\mathbf{t}}}
\def\rvu{{\mathbf{u}}}
\def\rvv{{\mathbf{v}}}
\def\rvw{{\mathbf{w}}}
\def\rvx{{\mathbf{x}}}
\def\rvy{{\mathbf{y}}}
\def\rvz{{\mathbf{z}}}

\def\erva{{\textnormal{a}}}
\def\ervb{{\textnormal{b}}}
\def\ervc{{\textnormal{c}}}
\def\ervd{{\textnormal{d}}}
\def\erve{{\textnormal{e}}}
\def\ervf{{\textnormal{f}}}
\def\ervg{{\textnormal{g}}}
\def\ervh{{\textnormal{h}}}
\def\ervi{{\textnormal{i}}}
\def\ervj{{\textnormal{j}}}
\def\ervk{{\textnormal{k}}}
\def\ervl{{\textnormal{l}}}
\def\ervm{{\textnormal{m}}}
\def\ervn{{\textnormal{n}}}
\def\ervo{{\textnormal{o}}}
\def\ervp{{\textnormal{p}}}
\def\ervq{{\textnormal{q}}}
\def\ervr{{\textnormal{r}}}
\def\ervs{{\textnormal{s}}}
\def\ervt{{\textnormal{t}}}
\def\ervu{{\textnormal{u}}}
\def\ervv{{\textnormal{v}}}
\def\ervw{{\textnormal{w}}}
\def\ervx{{\textnormal{x}}}
\def\ervy{{\textnormal{y}}}
\def\ervz{{\textnormal{z}}}

\def\rmA{{\mathbf{A}}}
\def\rmB{{\mathbf{B}}}
\def\rmC{{\mathbf{C}}}
\def\rmD{{\mathbf{D}}}
\def\rmE{{\mathbf{E}}}
\def\rmF{{\mathbf{F}}}
\def\rmG{{\mathbf{G}}}
\def\rmH{{\mathbf{H}}}
\def\rmI{{\mathbf{I}}}
\def\rmJ{{\mathbf{J}}}
\def\rmK{{\mathbf{K}}}
\def\rmL{{\mathbf{L}}}
\def\rmM{{\mathbf{M}}}
\def\rmN{{\mathbf{N}}}
\def\rmO{{\mathbf{O}}}
\def\rmP{{\mathbf{P}}}
\def\rmQ{{\mathbf{Q}}}
\def\rmR{{\mathbf{R}}}
\def\rmS{{\mathbf{S}}}
\def\rmT{{\mathbf{T}}}
\def\rmU{{\mathbf{U}}}
\def\rmV{{\mathbf{V}}}
\def\rmW{{\mathbf{W}}}
\def\rmX{{\mathbf{X}}}
\def\rmY{{\mathbf{Y}}}
\def\rmZ{{\mathbf{Z}}}

\def\ermA{{\textnormal{A}}}
\def\ermB{{\textnormal{B}}}
\def\ermC{{\textnormal{C}}}
\def\ermD{{\textnormal{D}}}
\def\ermE{{\textnormal{E}}}
\def\ermF{{\textnormal{F}}}
\def\ermG{{\textnormal{G}}}
\def\ermH{{\textnormal{H}}}
\def\ermI{{\textnormal{I}}}
\def\ermJ{{\textnormal{J}}}
\def\ermK{{\textnormal{K}}}
\def\ermL{{\textnormal{L}}}
\def\ermM{{\textnormal{M}}}
\def\ermN{{\textnormal{N}}}
\def\ermO{{\textnormal{O}}}
\def\ermP{{\textnormal{P}}}
\def\ermQ{{\textnormal{Q}}}
\def\ermR{{\textnormal{R}}}
\def\ermS{{\textnormal{S}}}
\def\ermT{{\textnormal{T}}}
\def\ermU{{\textnormal{U}}}
\def\ermV{{\textnormal{V}}}
\def\ermW{{\textnormal{W}}}
\def\ermX{{\textnormal{X}}}
\def\ermY{{\textnormal{Y}}}
\def\ermZ{{\textnormal{Z}}}

\def\vzero{{\bm{0}}}
\def\vone{{\bm{1}}}
\def\vmu{{\bm{\mu}}}
\def\vtheta{{\bm{\theta}}}
\def\va{{\bm{a}}}
\def\vb{{\bm{b}}}
\def\vc{{\bm{c}}}
\def\vd{{\bm{d}}}
\def\ve{{\bm{e}}}
\def\vf{{\bm{f}}}
\def\vg{{\bm{g}}}
\def\vh{{\bm{h}}}
\def\vi{{\bm{i}}}
\def\vj{{\bm{j}}}
\def\vk{{\bm{k}}}
\def\vl{{\bm{l}}}
\def\vm{{\bm{m}}}
\def\vn{{\bm{n}}}
\def\vo{{\bm{o}}}
\def\vp{{\bm{p}}}
\def\vq{{\bm{q}}}
\def\vr{{\bm{r}}}
\def\vs{{\bm{s}}}
\def\vt{{\bm{t}}}
\def\vu{{\bm{u}}}
\def\vv{{\bm{v}}}
\def\vw{{\bm{w}}}
\def\vx{{\bm{x}}}
\def\veps{{\bm{\epsilon}}}
\def\vtx{\widetilde{\bm{x}}}
\def\vty{\widetilde{\bm{y}}}
\def\vy{{\bm{y}}}
\def\vz{{\bm{z}}}

\def\stdnormal{\mathcal{N}(\bm{0}, \bm{I})}

\def\evalpha{{\alpha}}
\def\evbeta{{\beta}}
\def\evepsilon{{\epsilon}}
\def\evlambda{{\lambda}}
\def\evomega{{\omega}}
\def\evmu{{\mu}}
\def\evpsi{{\psi}}
\def\evsigma{{\sigma}}
\def\evtheta{{\theta}}
\def\eva{{a}}
\def\evb{{b}}
\def\evc{{c}}
\def\evd{{d}}
\def\eve{{e}}
\def\evf{{f}}
\def\evg{{g}}
\def\evh{{h}}
\def\evi{{i}}
\def\evj{{j}}
\def\evk{{k}}
\def\evl{{l}}
\def\evm{{m}}
\def\evn{{n}}
\def\evo{{o}}
\def\evp{{p}}
\def\evq{{q}}
\def\evr{{r}}
\def\evs{{s}}
\def\evt{{t}}
\def\evu{{u}}
\def\evv{{v}}
\def\evw{{w}}
\def\evx{{x}}
\def\evy{{y}}
\def\evz{{z}}

\def\mA{{\bm{A}}}
\def\mB{{\bm{B}}}
\def\mC{{\bm{C}}}
\def\mD{{\bm{D}}}
\def\mE{{\bm{E}}}
\def\mF{{\bm{F}}}
\def\mG{{\bm{G}}}
\def\mH{{\bm{H}}}
\def\mI{{\bm{I}}}
\def\mJ{{\bm{J}}}
\def\mK{{\bm{K}}}
\def\mL{{\bm{L}}}
\def\mM{{\bm{M}}}
\def\mN{{\bm{N}}}
\def\mO{{\bm{O}}}
\def\mP{{\bm{P}}}
\def\mQ{{\bm{Q}}}
\def\mR{{\bm{R}}}
\def\mS{{\bm{S}}}
\def\mT{{\bm{T}}}
\def\mU{{\bm{U}}}
\def\mV{{\bm{V}}}
\def\mW{{\bm{W}}}
\def\mX{{\bm{X}}}
\def\mY{{\bm{Y}}}
\def\mZ{{\bm{Z}}}
\def\mBeta{{\bm{\beta}}}
\def\mPhi{{\bm{\Phi}}}
\def\mLambda{{\bm{\Lambda}}}
\def\mSigma{{\bm{\Sigma}}}

\newcommand{\tens}[1]{\bm{\mathsfit{#1}}}
\def\tA{{\tens{A}}}
\def\tB{{\tens{B}}}
\def\tC{{\tens{C}}}
\def\tD{{\tens{D}}}
\def\tE{{\tens{E}}}
\def\tF{{\tens{F}}}
\def\tG{{\tens{G}}}
\def\tH{{\tens{H}}}
\def\tI{{\tens{I}}}
\def\tJ{{\tens{J}}}
\def\tK{{\tens{K}}}
\def\tL{{\tens{L}}}
\def\tM{{\tens{M}}}
\def\tN{{\tens{N}}}
\def\tO{{\tens{O}}}
\def\tP{{\tens{P}}}
\def\tQ{{\tens{Q}}}
\def\tR{{\tens{R}}}
\def\tS{{\tens{S}}}
\def\tT{{\tens{T}}}
\def\tU{{\tens{U}}}
\def\tV{{\tens{V}}}
\def\tW{{\tens{W}}}
\def\tX{{\tens{X}}}
\def\tY{{\tens{Y}}}
\def\tZ{{\tens{Z}}}

\def\gA{{\mathcal{A}}}
\def\gB{{\mathcal{B}}}
\def\gC{{\mathcal{C}}}
\def\gD{{\mathcal{D}}}
\def\gE{{\mathcal{E}}}
\def\gF{{\mathcal{F}}}
\def\gG{{\mathcal{G}}}
\def\gH{{\mathcal{H}}}
\def\gI{{\mathcal{I}}}
\def\gJ{{\mathcal{J}}}
\def\gK{{\mathcal{K}}}
\def\gL{{\mathcal{L}}}
\def\gM{{\mathcal{M}}}
\def\gN{{\mathcal{N}}}
\def\gO{{\mathcal{O}}}
\def\gP{{\mathcal{P}}}
\def\gQ{{\mathcal{Q}}}
\def\gR{{\mathcal{R}}}
\def\gS{{\mathcal{S}}}
\def\gT{{\mathcal{T}}}
\def\gU{{\mathcal{U}}}
\def\gV{{\mathcal{V}}}
\def\gW{{\mathcal{W}}}
\def\gX{{\mathcal{X}}}
\def\gY{{\mathcal{Y}}}
\def\gZ{{\mathcal{Z}}}

\def\sA{{\mathbb{A}}}
\def\sB{{\mathbb{B}}}
\def\sC{{\mathbb{C}}}
\def\sD{{\mathbb{D}}}
\def\sF{{\mathbb{F}}}
\def\sG{{\mathbb{G}}}
\def\sH{{\mathbb{H}}}
\def\sI{{\mathbb{I}}}
\def\sJ{{\mathbb{J}}}
\def\sK{{\mathbb{K}}}
\def\sL{{\mathbb{L}}}
\def\sM{{\mathbb{M}}}
\def\sN{{\mathbb{N}}}
\def\sO{{\mathbb{O}}}
\def\sP{{\mathbb{P}}}
\def\sQ{{\mathbb{Q}}}
\def\sR{{\mathbb{R}}}
\def\sS{{\mathbb{S}}}
\def\sT{{\mathbb{T}}}
\def\sU{{\mathbb{U}}}
\def\sV{{\mathbb{V}}}
\def\sW{{\mathbb{W}}}
\def\sX{{\mathbb{X}}}
\def\sY{{\mathbb{Y}}}
\def\sZ{{\mathbb{Z}}}

\def\R{{\mathbb{R}}}

\def\emLambda{{\Lambda}}
\def\emA{{A}}
\def\emB{{B}}
\def\emC{{C}}
\def\emD{{D}}
\def\emE{{E}}
\def\emF{{F}}
\def\emG{{G}}
\def\emH{{H}}
\def\emI{{I}}
\def\emJ{{J}}
\def\emK{{K}}
\def\emL{{L}}
\def\emM{{M}}
\def\emN{{N}}
\def\emO{{O}}
\def\emP{{P}}
\def\emQ{{Q}}
\def\emR{{R}}
\def\emS{{S}}
\def\emT{{T}}
\def\emU{{U}}
\def\emV{{V}}
\def\emW{{W}}
\def\emX{{X}}
\def\emY{{Y}}
\def\emZ{{Z}}
\def\emSigma{{\Sigma}}

\newcommand{\etens}[1]{\mathsfit{#1}}
\def\etLambda{{\etens{\Lambda}}}
\def\etA{{\etens{A}}}
\def\etB{{\etens{B}}}
\def\etC{{\etens{C}}}
\def\etD{{\etens{D}}}
\def\etE{{\etens{E}}}
\def\etF{{\etens{F}}}
\def\etG{{\etens{G}}}
\def\etH{{\etens{H}}}
\def\etI{{\etens{I}}}
\def\etJ{{\etens{J}}}
\def\etK{{\etens{K}}}
\def\etL{{\etens{L}}}
\def\etM{{\etens{M}}}
\def\etN{{\etens{N}}}
\def\etO{{\etens{O}}}
\def\etP{{\etens{P}}}
\def\etQ{{\etens{Q}}}
\def\etR{{\etens{R}}}
\def\etS{{\etens{S}}}
\def\etT{{\etens{T}}}
\def\etU{{\etens{U}}}
\def\etV{{\etens{V}}}
\def\etW{{\etens{W}}}
\def\etX{{\etens{X}}}
\def\etY{{\etens{Y}}}
\def\etZ{{\etens{Z}}}

\newcommand{\pdata}{p_{\rm{data}}}
\newcommand{\ptrain}{\hat{p}_{\rm{data}}}
\newcommand{\Ptrain}{\hat{P}_{\rm{data}}}
\newcommand{\pmodel}{p_{\rm{model}}}
\newcommand{\Pmodel}{P_{\rm{model}}}
\newcommand{\ptildemodel}{\tilde{p}_{\rm{model}}}
\newcommand{\pencode}{p_{\rm{encoder}}}
\newcommand{\pdecode}{p_{\rm{decoder}}}
\newcommand{\precons}{p_{\rm{reconstruct}}}
\newcommand{\one}[1]{\mathbbm{1}{[#1]}}
\newcommand{\laplace}{\mathrm{Laplace}} 

\newcommand{\E}{\mathbb{E}}
\newcommand{\Ls}{\mathcal{L}}
\renewcommand{\R}{\mathbb{R}}
\newcommand{\emp}{\tilde{p}}
\newcommand{\lr}{\alpha}
\newcommand{\reg}{\lambda}
\newcommand{\rect}{\mathrm{rectifier}}
\newcommand{\softmax}{\mathrm{softmax}}
\newcommand{\sigmoid}{\sigma}
\newcommand{\softplus}{\zeta}
\newcommand{\KL}{D_{\mathrm{KL}}}
\newcommand{\Var}{\mathrm{Var}}
\newcommand{\standarderror}{\mathrm{SE}}
\newcommand{\Cov}{\mathrm{Cov}}
\newcommand{\normlzero}{L^0}
\newcommand{\normlone}{L^1}
\newcommand{\normltwo}{L^2}
\newcommand{\normlp}{L^p}
\newcommand{\normmax}{L^\infty}

\newcommand{\parents}{Pa} 

\let\ab\allowbreak


\section{Introduction}
\label{sec:intro}

The buildings of a city are a pivotal element that molds its urban structure and morphology, which serve various functions, encompassing commerce, residential areas, and industrial zones. Understanding these functions proves instrumental in tasks such as map generalization, delineating urban zones, deciphering land use patterns\citep{land11070976}, and aiding governmental management. Furthermore, the classification of building functions holds immense significance across diverse applications, spanning from assessing energy demands, urban climate studies, and energy balance modeling \citep{NathalieTornay2017GENIUSAM} to conducting analyses of urban social dynamics \citep{doi:10.2747/1548-1603.42.1.80}. Consequently, the accurate and fine classification of buildings on a urban scale has emerged as a focal topic within the research field of urban remote sensing .

Buildings are often the basic units for cartography or urban planning on vector maps, and learning the function of a building significantly impacts urban transportation and resource management. However, local authorities or national mapping agencies sometimes record the function information of a building, and such data are usually not publicly available \citep{CidliaCostaFonte2018ClassificationOB}. The widely used commercial map servers, such as Google Maps and Baidu Maps, can only provide points of interest rather than the function of buildings; thus, functional buildings are unavailable through commercial map servers.

In recent years, advanced remote sensing image analysis methods have been developed, especially for very high-resolution satellite images, and used for information extraction, thanks to high information details and wide availability \citep{yao2009,7378854,YongyangXu2018BuildingEI,9027926,POLEWSKI2021297,YongyangXu2022BuildingFC}. Some studies have started to focus on classifying building types based on spectral characteristics extracted from remote sensing images. To identify a specific type of buildings, the role of spatial, structural, and contextual features, including gray-level co-occurrence matrices, histograms of oriented gradients and line support regions have been analyzed \citep{JGraesser2012ImageBC}. Then, defining urban neighborhoods as homogeneous zones, and classifying them as formal and informal areas. Moreover, pixel-based classification methods have been applied to satellite images to extract spectral information for characterising roof types and consecutively building types \citep{HannesTaubenbck2009AssessingBV}. Mathematical morphology have also been deployed for building function classification. While earth observation data are widely used for the extraction of multi-scale, area-wide information on general urban structure, the derivation of fine building types remains a challenging and difficult task. Former methods could achieve a building function classification scheme, which treats remote sensing image pixels as the spatial entity for building function classification; the geometric information of buildings may be ignored, such as edge or corner information. As a result, classification results cannot serve as a precise base map for cartography or city planning. Therefore, we need to develop new techniques for analyzing instance-level urban building functions.

With the fast development of artificial intelligence, deep learning and machine learning methods have been widely applied to building type classification \citep{SaraShirowzhan2017BuildingCF}. For instance, \cite{ChristophRmer2010IdentifyingAS} and \cite{HennAndr2012AutomaticCO} analyzed the architectural building type (detached building, semi-detached building, terraced building, villa, Wilhelminian-style building, etc.) from very coarse 3D city model data based on support vector machines (SVMs). As convolutional neural networks have been widely developed in computer version, some neural networks have been designed for the building type classification by analyzing street view images. \cite{EikeJensHoffmann2019ModelFF} proposed a fusion model for building type classification from aerial and street view images; Google Street View images were also used for multi-label building function classification using convolutional neural networks \citep{KANG201844,ShivangiSrivastava2018MultilabelBF}. \cite{SalmaTaoufiq2020HierarchyNetHC} proposed a new hierarchical network, named as HierarchyNet, for classifying urban buildings across the globe into different main and subcategories using facade images. Moreover,  only roadside buildings are easy to be observed and can be acquired in the street view images. Therefore, a new and more generalizable satellite remote sensing based method is required for large-scale fine-grained building function classification.

Building footprints are useful for a range of important applications, from population estimation, urban planning and humanitarian response, to environmental and climate science. Google released Open Buildings\footnote{\url{https://sites.research.google/open-buildings/}} based on previous work~\citep{sirko2021continental}. Open Buildings is a large-scale open dataset which contains 1.8 billion building outlines derived from high-resolution satellite imagery all around the world. For each building in this dataset, a polygon describing its footprint on the ground and a plus code corresponding to the centre of the building are recorded. There is no information about the type of building, its street address, or any details other than its boundary geometry and geolocation. Microsoft also released 1.28 billion building footprints and 174 million building height around the world estimated from Bing Maps imagery between 2014 and 2023\footnote{\url{https://github.com/microsoft/GlobalMLBuildingFootprints}}, the data set is freely available for download. Previous studies can broadly categorise land use based on footprint and satellite imagery, but not able to provide a fine-grained categorisation of building types.

Buildings of various functions exhibit different features, such as industrial buildings always have larger footprint areas than residential buildings, whereas official buildings are higher than industrial buildings. The function of urban buildings is strongly correlated with environmental and social variables \citep{ShihongDu2015SemanticCO}. Moreover, the building function always has certain spatial relations with their neighbors. For example, residential buildings are always regularly co-spaced with each other, and industrial buildings are located far away from residential buildings. 


To maintain the geometry information during classification of the building function types, shape-based methods using building footprints have been proposed by researchers. However, they offer the ability to incorporate shape-based features such as building geometry and morphology for building type classification, including 1D features, such as length, width, and length–width ratios \citep{HennAndr2012AutomaticCO}, 2D features such as area \citep{PatrickLscher2009IntegratingOM}, building elongation, compactness, rectangularity, and topological features such as the number of vertices \citep{StefanSteiniger2008AnAF}, 3D features such as building height, and the number of stories \citep{DavidHa2017ANR, HennAndr2012AutomaticCO}. In these methods, individual building polygons are treated as the spatial entity for building function classification. Although the geometric information of a building boundary can be obtained, the descriptors can hardly retain the complete building geometry due to lack of image texture information. To the best of our knowledge, the research work presented in this paper is the first attempt to develop satellite imagery based solution to fine-grained building instance classification in dense urban areas.


%
To sum up, the main contributions of this paper are concluded as follows:
\begin{itemize}
\item We propose a pioneering framework for the fine classification of buildings in a dense urban area solely utilizing low-resolution overhead images, such as Google Earth satellite images. The approach incorporates an innovative Diffusion Probabilistic super-resolution module for enhancing the image quality, which is strategically designed to bridge the domain-specific knowledge gap. 

\item We introduce a category information balancing module known as CIBM, which plays a pivotal role in rectifying class imbalances by dynamically regulating the inclusion of images from different categories. The CIBM not only enhances the model robustness but also fosters equivalent performance across diverse classes.

\item Our methodology is comprehensively validated through a series of comparative and ablation experiments. The outcomes unequivocally underscore the efficacy of our proposed approach. Although our ultimate experimental results may not attain absolute perfection, the method establishes a level of classification accuracy that is on par with street-view image based techniques, despite relying solely on low-resolution satellite images.
\end{itemize}


\section{Related works}

\subsection{\textbf{Building Classification with Street View Images}}
\cite{laupheimer2018neural} categorized terrestrial images of building facades into five broad categories. They used Convolutional Neural Networks (CNNs) to classify the street view images. However, the error rate of 36\% misclassified images highlights the necessity for further improvement. ~\cite{KANG201844} obtained Google Street View images from the USA and Canada to perform architectural semantic classification using CNN, instead of  directly using the satellite imagery. Recognizing buildings from street-view images and encoding them for image classification, as proposed by ~\cite{9380541}, is also a ground-breaking but useful approach. Recently, some researchers classified urban buildings into 10 fine categories using graph neural networks based on topology, achieving an accuracy of Top-1 46.2\%, Top-5 82.4\% \citep{ZHANG2023153}. These methods  offer useful guidance for the functional classification of urban buildings  using streetscape images. However, they do have some limitations. First, street view images are high-resolution images, which can be costly to obtain and may result in omission errors when buildings are obstructed. So, not all buildings can be classified using this approach. Secondly, it is impossible and inefficient for street view image to perform a city-scale building categorisation, since the data collection is very costly and limited to buildings in the vicinity of transportation network.


\subsection{\textbf{Building Classification with Satellite Images}}
~\cite{xiao2020efficient} proposed the utilization of oblique-view images to categorize building functions. which classified the building functions into four distinct categories during experimentation. Subsequently, the final test demonstrated a classification accuracy of 60\%~\citep{xiao2020efficient}. ~\cite{HuangXingliang9857458,HuangXingliang10237272} conducted a study on building detection and classification using very high resolution satellite images with a GSD of 0.5-0.8m. The focus was on the object-level interpretation of individual buildings, enabling a 5-category vocabulary classification of buildings. However, the work required extensive use of densely pre-labeled semantic information, which is known to be very labour-intensive. Similarly, the method does not export accurate category boundaries for individual buildings, especially when several buildings of the same category are located in close proximity. 

\subsection{\textbf{Satellite Image Super-resolution}}
Numerous super-resolution methods have been proposed in the computer vision community \citep{ahn-cvpr-2018,ledig2017photo, sajjadi2017enhancenet}. Many of early works on super-resolution is based on regression and trained with an MSE loss \citep{ahn-cvpr-2018, KimCVPR2016}. Auto-regressive models have been successfully used for super-resolution and cascaded up-sampling \citep{menick-iclr-2019,parmar2018image}. However, due to the inherent complexity of real-world remote sensing images, current models are prone to color distortion, blurred edges, and unrealistic artifacts, making it difficult to adapt these methods from ordinary images to satellite images by considering the distinct domain shift. ~\cite{rs15051391} proposed a second-order attention generator adversarial attention network (SA-GAN) model to address existing problems. ~\cite{9861276} proposed an arbitrary scale SR network for satellite image reconstruction, enhancing the high-frequency details in satellite images with the help of edge reinforcement module. However, these methods can not achieve fine control of the super-resolution process.


\subsection{\textbf{Category Balancing Problem in Image Classification}}
~\cite{vannucci2016smart} proposed a radial basis-based under-sampling technique: removing commonly occurring samples in the training set and adaptively determining the optimal imbalance rate for various datasets. This technique resulted in improved model classification performance and enhanced model generalisation abilities. ~\cite{hasib2021hsdlm} proposed the Hybrid Sampling with Deep Learning Method (HSDLM). The dataset is pre-processed via label coding, with noise being removed through the under-sampling algorithm. They also use the SMOTE over-sampling technique to balance the data and implements three parallel types of LSTM to improve the accuracy. These works aim to address category imbalance problem through methods such as up-sampling and down-sampling. However, none of these approaches take into account the issue of intra-category sample similarity.

\section{Methodology}
\label{sec:method}

\subsection{\textbf{Overview}}
\label{sec:overview}

\begin{figure*}[t]
\begin{center}
\includegraphics[width=0.95\textwidth]{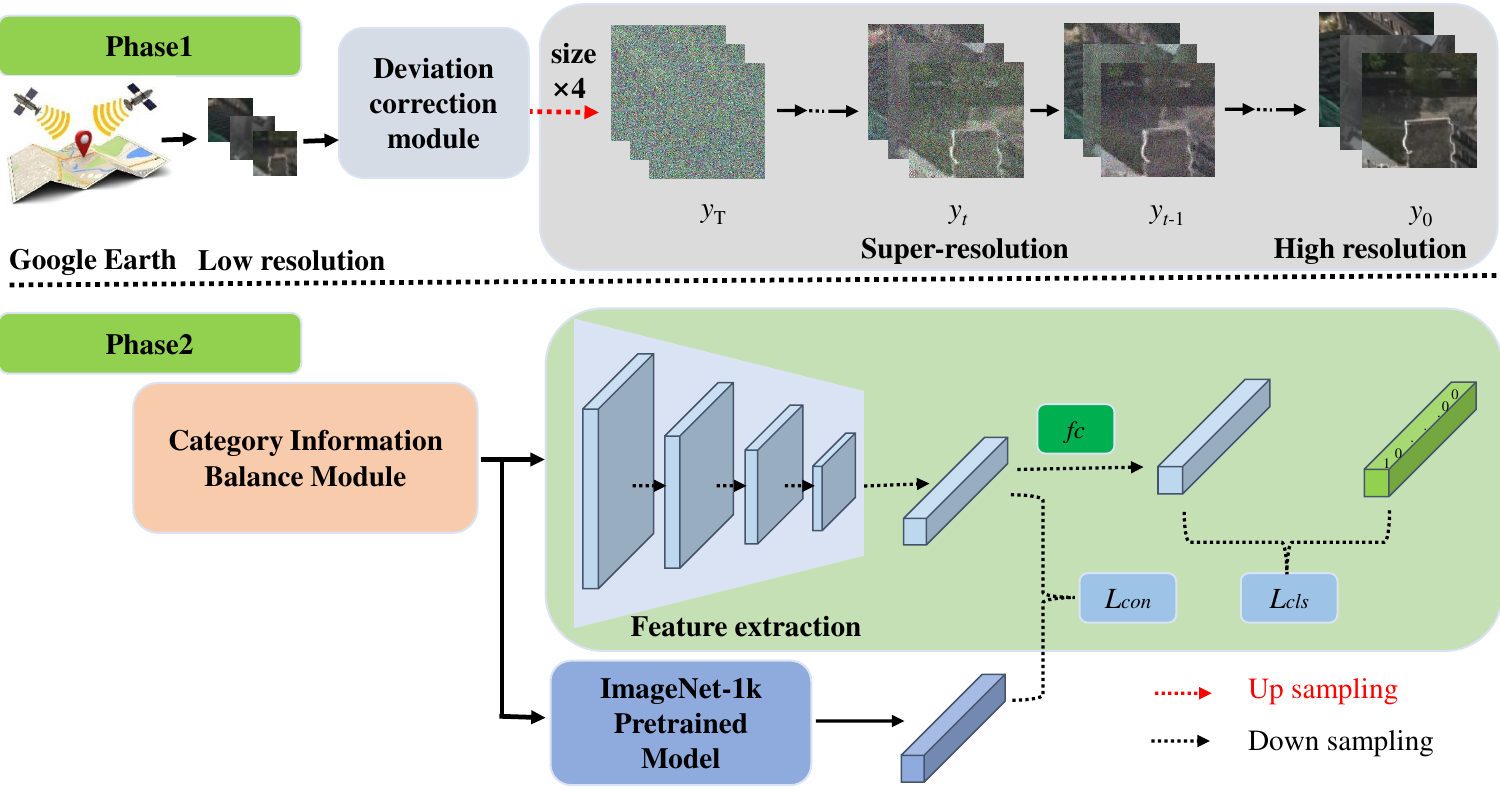}
\end{center}
\caption{Overview of the proposed building category classification network based on Google Earth satellite imagery. }
\label{fig:overview}
\end{figure*}

In this section we describe how our approach works to solve problems mentioned above. As shown in the Fig.\ref{fig:overview}, the processing flow is divided into two phases. The first phase is a super-resolution network for low-resolution Google Earth satellite images based on Denoising Diffusion Probabilistic Model(DDPM). In this part, we proposed a deviation correction module to mitigate the impacts of feature discrepancy between the aerial photography and Google Earth satellite images, as we trained a DDPM model with low-resolution(LR) and high-resolution(HR) aerial image pairs. Maintaining congruence in both resolution and size between the input training and the target images to be super-resolved ensures that the network skilfully captures the inherent features in satellite imagery. This approach warrants that intricate details are not unduly distorted in the resulting super-resolution images. With this SR model, target satellite LR images are transformed to HR images and the problem of lacking metre-scale building details and features is alleviated ~\citep{Saharia2023_SR3}.

The second phase is the proposed Urban Building Fine-grained Classification Network (UB-FineNet). Taking efficiency and lightweight into consideration, we adapted backbone from the ShufffleNetV2 ~\citep{ma2018shufflenetv2}, and proposed a category information balanced module to alleviate the imbalanced category information and to improve robustness of UB-FineNet. Then, in the proposed contrastive learning strategy, the UB-FineNet is supervised by output features of existing models trained on the ImageNet-1k dataset ~\citep{russakovsky2015imagenet1k}, which is very easily accesssiable. In this process, the knowledge of the existing models is distilled and passed on to the newly trained model, which improved the performance and convergence speed of our network. 

\subsection{\textbf{Image Super-Resolution}}
\label{sec:method_srDDPM}

\subsubsection{\textbf{Conditional Denoising Diffusion Model}}
\label{sec:method_srDDPM12}
Assume that there is a dataset $\mathcal{D_{0}} = \{\vl_i, \vy_i\}_{i=1}^N$, which contains LR-HR image pairs. We rescale LR images $\vl_i$ by interpolation to the same size as HR images $\vy_i$, denoted as $\vx_i$. Then we get the new image pairs dataset, denoted as $\mathcal{D} = \{\vx_i, \vy_i\}_{i=1}^N$. Given source image $\vx_i$, we hope to obtain the corresponding target HR image $\vy_i$. However, the conditional distribution $p(\vy \,|\, \vx)$ is unknown, leading to confusing returns. We want to solve this problem by adapting the conditional DDPM model~\citep{ho2020denoising,Saharia2023_SR3} , building a new network whose parameters can be learned and optimised in stochastic iterations to estimate the  probabilities $p(\vy \,|\, \vx)$.

\begin{figure}[h]
\begin{center}
\includegraphics[width=0.5\textwidth]{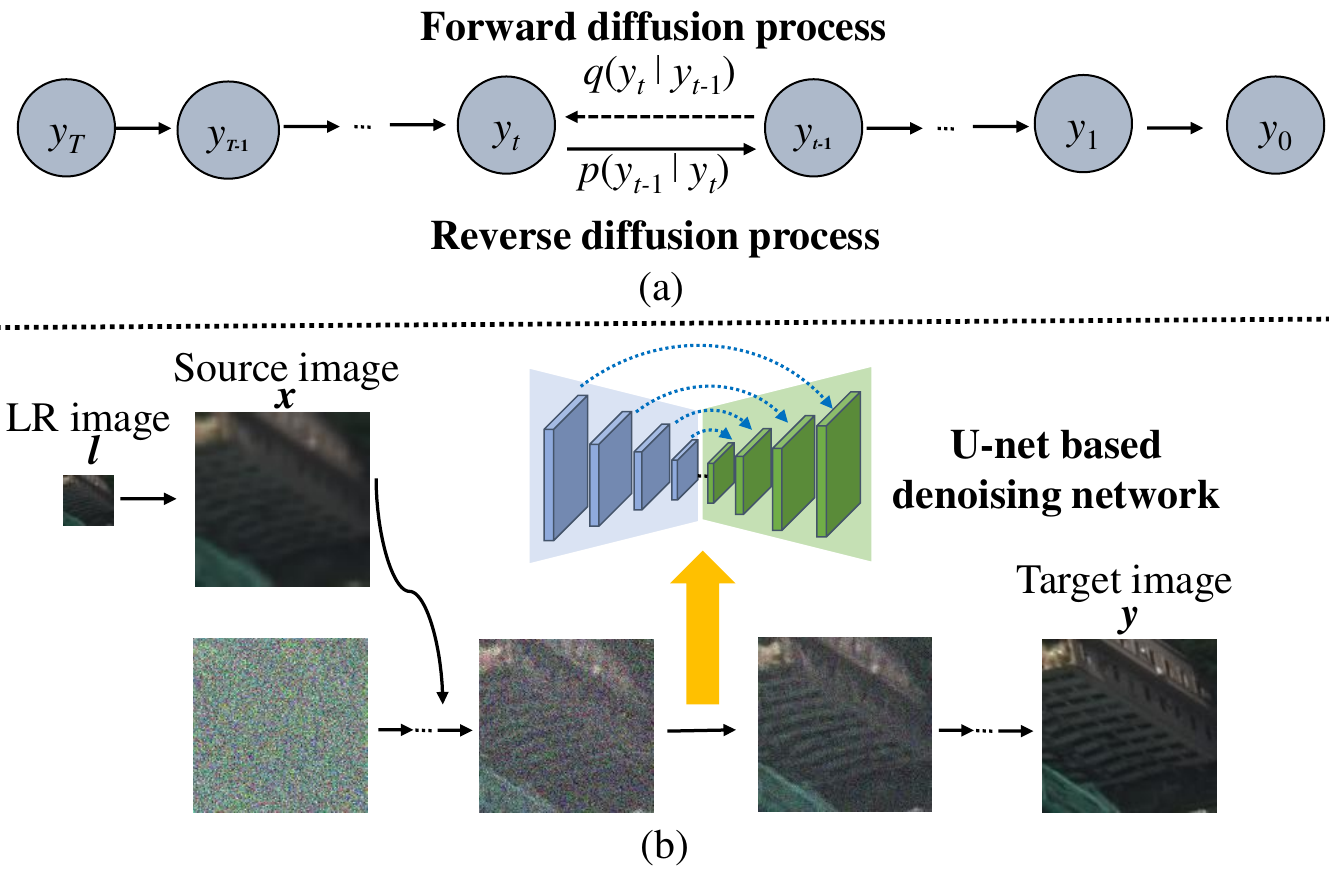}
\end{center}
\caption{Schematic representation of the architecture of denoising diffusion probabilistic model (DDPM). (a) The diffusion process indicates the gradual process adding Gaussian noise to the target image $\vy_{0}$ (from right to left), the reverse diffusion process depicts the gradual process of removing Gaussian noise from the source image $\vy_{T}$(from right to left). (b) Reverse diffusion process from low-resolution image with trainable U-net based denoising network.}
\label{fig:ddpm}
\end{figure}

Suppose that the low-resolution images are of poorer quality due to noises superimposed on high-resolution images, so we can use DDPM to denoise the low-resolution source image $\vl$ using the reverse diffusion process, thus obtaining a higher-resolution target image $\vy$. The process of generating a target HR image from the conditional DDPM is divided into $T$-steps, step by step, as shown in Fig.~\ref{fig:ddpm}. 
This process starts with a pure Gaussian noise image $\vy_T \sim \mathcal{N}(\bm{0}, \bm{I})$ and then iterates successively according to the conditional distribution  $p_\theta(\vy_{t-1} \,|\, \vy_t, \vx)$ learned by the network to obtain $\vy_{T-1}, \vy_{T-2}, \dotsc, \vy_{0}$ separately and all the steps are concatenated to achieve the generation of HR image $\vy_{0} \sim p(\vy \, |\, \vx)$.

\subsubsection{\textbf{Denoising Model Training}}
We also consider the image noise accumulation process as a Markov chain and the denoising process as an inverse process. To train the parameters of the denoising network $\mathcal{F}_\theta$~\citep{ho2020denoising,Saharia2023_SR3}, the given source image $\vx$ and the image $\vy^m$ generated by the intermediate process are input to the network. $\vy^m$ can be expressed as:
\begin{equation}
    \vy^m = \sqrt{\gamma}\, \vy_0 + \sqrt{1-\gamma} \,\veps~,
    ~~~~~~ \veps \sim \mathcal{N}(\bm{0},\bm{I})~
\label{eq:noisy-y}
\end{equation}
The denoising model $\mathcal{F}_\theta(\vx, \vy^m, \gamma)$ takes the source image, intermediate image and the statistics for the variance of Gaussian noise $\gamma$ as input, the network parameters are iteratively updated. The noise vector superimposed on the source image $\veps$  at each stage is estimated.

Following ~\cite{chen-iclr-2021} and ~\cite{Saharia2023_SR3}, we set a variable $\gamma$ and condition it so that the denoising model $\mathcal{F}_\theta$ can be well aware of noises. The objective function for training $\mathcal{F}_\theta$ can be expressed as
\begin{equation}
    \E_{(\vx, \vy)} \E_{\veps, \gamma} \bigg\lVert \mathcal{F}_\theta(\vx, \sqrt{\gamma} \,\vy_0 + \sqrt{1-\gamma}\, \veps, \gamma) - \veps\, \bigg\rVert^{a}_a~
\label{eq:loss}
\end{equation}
where $\veps \sim \mathcal{N}(\bm{0}, \bm{I})$, $(\vx, \vy)$ is selected image pairs from the training dataset, variant $a \in \{1, 2\}$, which means the sum of L1 Norm and squares of L2 Norm, and $\gamma \sim p(\gamma)$.

\algrenewcommand\algorithmicindent{0.5em}%
\begin{figure}[t]
\vspace*{-.5cm}
\small
\begin{minipage}[t]{0.45\textwidth}
\begin{algorithm}[H]
  \caption{Train a denoising model $\mathcal{F}_\theta$} 
  \label{alg:ddpm training}
  \small
  \begin{algorithmic}[1]
    \Repeat
      \State $(\vx, \vy_0) \sim p(\vx, \vy)$
      \State $t \sim \mathrm{Uniform}(\{1, \dotsc, T\})$
      \State $\gamma \sim p(\gamma)$
      \State $\bm{\epsilon}\sim\mathcal{N}(\mathbf{0},\mathbf{I})$
      \State Gradient descent
      \Statex $\qquad \nabla_\theta \left\lVert \mathcal{F}_\theta(\vx, \sqrt{\gamma} \vy_0 + \sqrt{1-\gamma} \bm{\epsilon}, \gamma) - \veps \right\rVert_a^a$ 
    \Until{Converged}
  \end{algorithmic}
\end{algorithm}
\end{minipage}
\vspace*{-.2cm}
\end{figure}

As shown in Algorithm~\ref{alg:ddpm training} and Eq.~\eqref{eq:loss}, we can compute the output of $\mathcal{F}_\theta$ step by step until the target image $\vy_0$ is generated. Given $\gamma$ and $\vy^m$, $\veps~$ can be estimated from the original image $\vy_0$ deterministically, vice versa.

\subsubsection{\textbf{Deviation Correction Module}}

\begin{figure}[t]
\begin{center}
\includegraphics[width=0.5\textwidth]{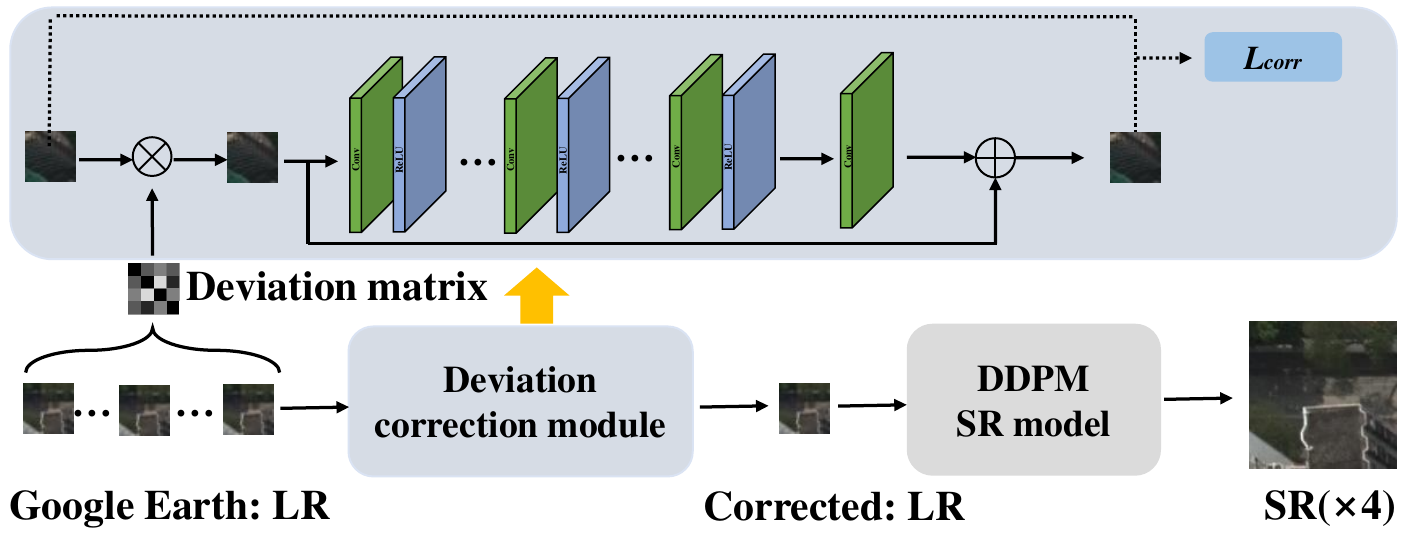}
\end{center}
\caption{Deviation correction module.}
\label{fig:Deviation correction module}
\end{figure}

The training of the super-resolution network is supervised by HR images, so the overall feature distribution and the detailed features of the training data can affect the network performance directly. If the domain of the inference and training images is not identical, i.e. a domain shift exists, the deviation needs to be corrected to avoid distorting features in inference results, as shown in Fig.\ref{fig:Deviation correction module}. In our model, the inference process is characterized as a reverse Markovian process, which operates against the direction of the forward diffusion process and starts from Gaussian noise $\mathit{y}_T$:

\begin{figure}
\vspace*{-.5cm}
\small
\begin{minipage}[t]{0.45\textwidth}
\begin{algorithm}[H]
  \caption{Inference in $T$ iterative refinement steps} 
  \label{alg:ddpm inference}
  \small
  \begin{algorithmic}[1]
    \vspace{.04in}
    \State $\vy_T \sim \mathcal{N}(\mathbf{0}, \mathbf{I})$
    \For{$t=T, \dotsc, 1$}
      \State $\vz \sim \mathcal{N}(\mathbf{0}, \mathbf{I})$ if $t > 1$, else $\vz = \mathbf{0}$
      \State $\vy_{t-1} = \frac{1}{\sqrt{\alpha_t}}\left( \vy_t - \frac{1-\alpha_t}{\sqrt{1-\gamma_t}} \mathcal{F}_\theta(\vx, \vy_t, \gamma_t) \right) + \sqrt{1 - \alpha_t} \vz$
    \EndFor
    \State \textbf{return} $\vy_0$
    \vspace{.04in}
  \end{algorithmic}
\end{algorithm}
\end{minipage}
\vspace*{-0.2cm}
\end{figure}

Inference under our model is defined as a \emph{reverse} Markovian process, which goes in the reverse direction of the forward diffusion process, starting from Gaussian noise $\mathit{y}_T$:

\begin{eqnarray}
    p_\theta(\mathit{y}_{0:T} | \vx) &=& p(\mathit{y}_T) \prod\nolimits_{t=1}^T p_\theta(\mathit{y}_{t-1} | \mathit{y}_t, \vx) \\
    p(\mathit{y}_T) &=& \mathcal{N}(\mathit{y}_T \mid \bm{0}, \bm{I}) \label{eq:mutheta}\\
    p_\theta(\mathit{y}_{t-1} | \mathit{y}_{t}, \vx) &=& \mathcal{N}(\mathit{y}_{t-1} \mid \mu_{\theta}(\vx, {\mathit{y}}_{t}, \gamma_t), \sigma_t^2\bm{I})~. \label{eq:reverse_process}
\end{eqnarray}
We define the inference process in terms of isotropic Gaussian conditional distributions, $p_\theta(\mathit{y}_{t-1} | \mathit{y}_{t}, \vx)$, which are learned.
If the noise variance of the forward process steps is minimized, \ie~$\alpha_{1:T} \approx 1$, the optimal reverse process $p(\mathit{y}_{t-1} | \mathit{y}_{t}, \vx)$ will be approximately Gaussian~\citep{sohl2015deep}.
Thus, the selection of Gaussian conditionals in the inference process Eq.\eqref{eq:reverse_process} can offer a satisfactory match to the actual reverse process.
Simultaneously, ensuring that $1 - \gamma_T$ is sufficiently large enables $\mathit{y}_T$ to be approximately distributed in line with the prior $p(\mathit{y}_T) = \mathcal{N}(\mathit{y}_T | \bm{0}, \bm{I})$, facilitating the sampling process to commence with pure Gaussian noises.

The denoising model $\mathcal{F}_\theta$ is trained to estimate noise parameters, given any intermediate images $\vy_m$ generated, which include $\mathit{y}^t$. Hence, with $\mathit{y}_t$ at hand, $\mathit{y}_0$ is estimated by reorganizing the terms in Eq.\eqref{eq:noisy-y} as following:
\begin{equation}
    \hat{\mathit{y}}_0 = \frac{1}{\sqrt{\gamma_t}} \left( \mathit{y}_t - \sqrt{1 - \gamma_t}\, \mathcal{F}_\theta(\vx, \mathit{y}_{t}, \gamma_t) \right)~
\end{equation}
Following the formulation of \cite{ho2020denoising,Saharia2023_SR3}, we substitute the estimate $\hat{\mathit{y}}_0$ into the posterior distribution of $q(\mathit{y}_{t-1} | \mathit{y}_0, \mathit{y}_t)$ to parameterize the mean of $p_\theta(\mathit{y}_{t-1} | \mathit{y}_t, \vx)$ as
\begin{equation}
    \mu_{\theta}(\vx, {\mathit{y}}_{t}, \gamma_t) = \frac{1}{\sqrt{\alpha_t}} \left( \mathit{y}_t - \frac{1-\alpha_t}{ \sqrt{1 - \gamma_t}} \mathcal{F}_\theta(\vx, \mathit{y}_{t}, \gamma_t) \right)~
\end{equation}
and set the variance of $p_\theta(\mathit{y}_{t-1}|\mathit{y}_t, \vx)$ to $(1 - \alpha_t)$, a default given by the variance of the forward process \citep{ho2020denoising,Saharia2023_SR3}.

Utilizing this parameterization, each step of iterative refinement within our model is structured as follows:

\begin{equation}
\mathit{y}_{t-1} \leftarrow \frac{1}{\sqrt{\alpha_t}} \left( \mathit{y}_t - \frac{1-\alpha_t}{ \sqrt{1 - \gamma_t}} \mathcal{F}_\theta(\vx, \mathit{y}_{t}, \gamma_t) \right) + \sqrt{1 - \alpha_t}\veps_t~
\end{equation}
where $\veps_t \sim \stdnormal$. This bears resemblance to a single step of Langevin dynamics, with $\mathcal{F}_\theta$ offering an approximation of the log-density's gradient of data.

\subsection{\textbf{Fine-grained Building Classification}}
\label{sec:CIBM_Building-ShufffleNet}


\subsubsection{\textbf{Category Information Balanced Module (CIBM)}}

\begin{figure}[t]
\begin{center}
\includegraphics[width=0.5\textwidth]{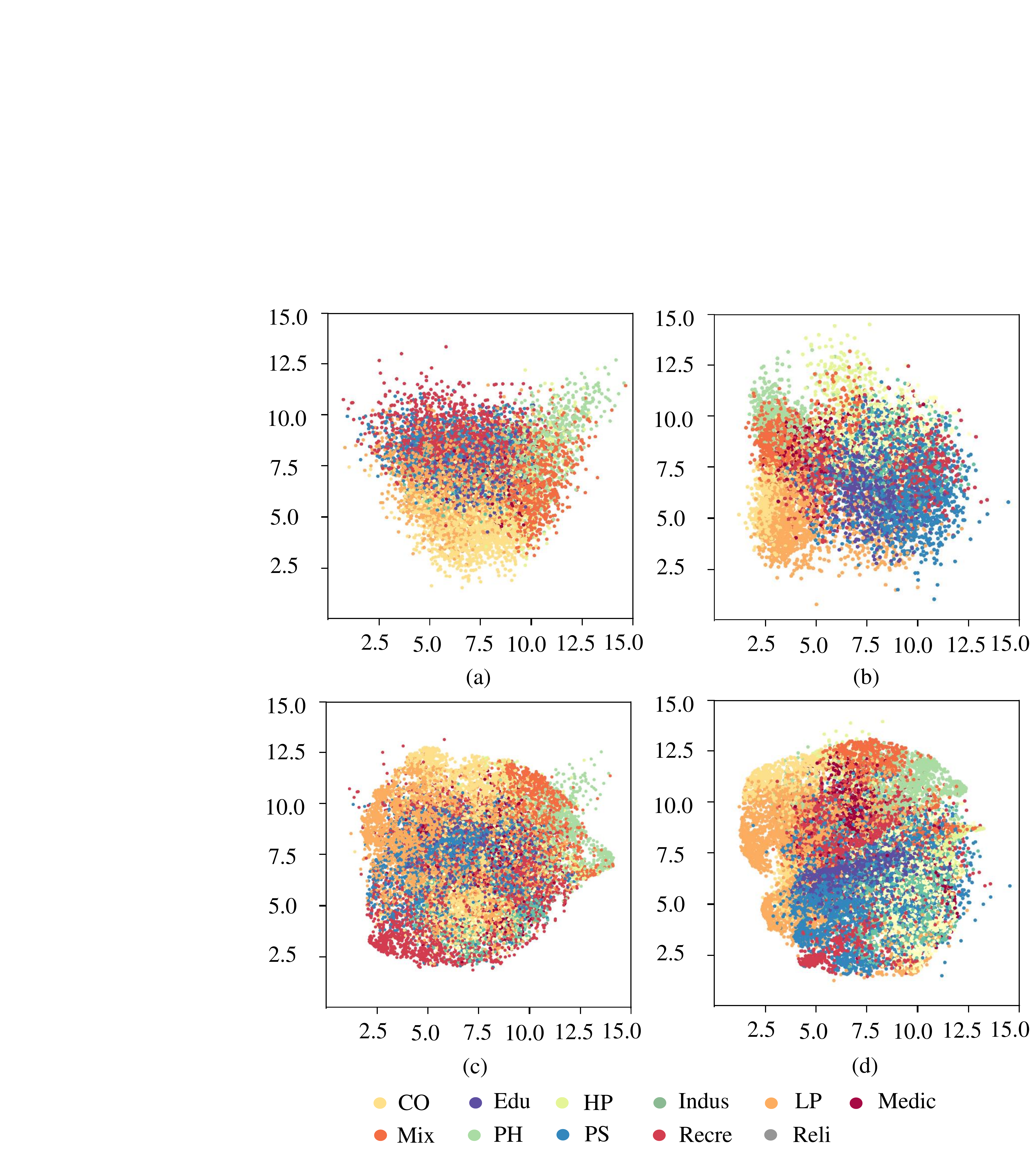}
\end{center}
\caption{Visualisation of features of different categories. (a) and (b) represent the intra-class distribution of spatial Euclidean distance of features produced by PCA before and after processing by the CIBM module, respectively, while (c) and (d) represent the intra-class distribution of spatial Euclidean distance of features produced by t-SNE before and after processing by the CIBM module, respectively.}

\label{fig:Visualisation of features}
\end{figure}

Category imbalance is one of the most important and common problem in image classification task. There are many reasons for this problem, such as the fact that data of certain category are more difficulty and costly to obtain than others or inherently less existent. In order to obtain the same results for a category with few samples as for those dominant categories, former studies have proposed many solutions to solve the widespread problem of category imbalance. Under-sampling is a common method to count the number of samples per category and calculate the weights for each category accordingly, with fewer samples receiving greater weights and vice versa, and adjust the number of samples per category input to the network by the weights for training. By reducing the number of categories with more data, it ensures that the input samples for each category are equivalent. The intra-class distribution of spatial Euclidean distance of features produced by PCA and t-SNE  before and after depolying CIBM module is shown in Fig.\ref{fig:Visualisation of features}, which shows that the features of similar samples after processing are more concentrated than before. Generative task networks, such as the Generative Adversarial Network (GAN), are another effective solution for this problem, where the number of samples input for the classification network is equalized for each category by supplementing the needy category with a small number of samples. Although all of these approaches have achieved improvement to varied degree, the category imbalance problem is only mitigated by equalizing the number of samples for each category, but not yet by considering the intra-class sample relevance. In this work, we proposed a new module which takes the feature relevance between samples within each of single categories into account. In the training phase, new weights are calculated and assigned to each category, which makes the training process more focused on the inter-class differences rather than purely on the number of samples, helping to improve the model robustness.

\begin{figure*}[t]
\begin{center}
\includegraphics[width=0.8\textwidth]{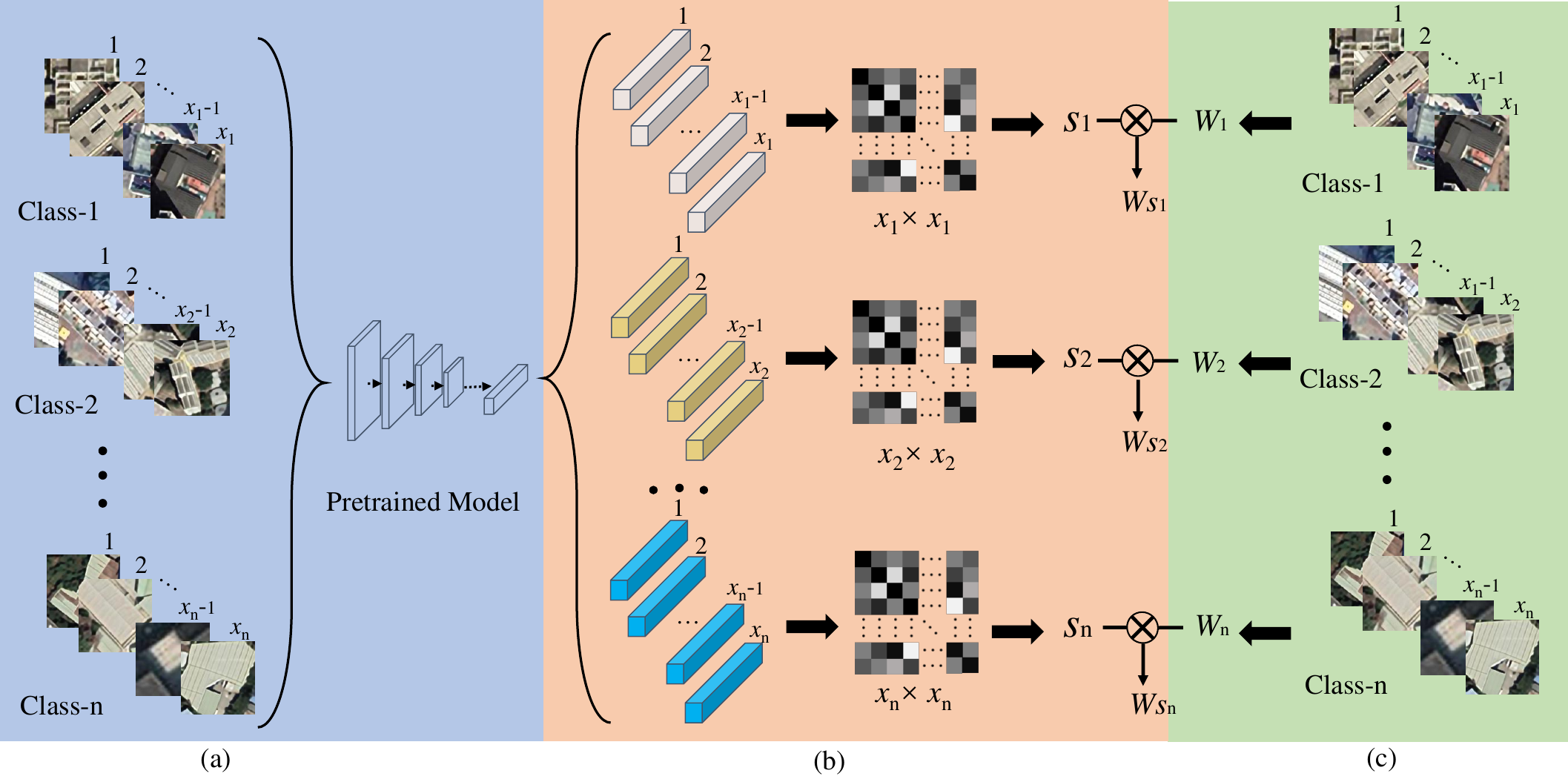}
\end{center}
\caption{Comparison of proposed category information balanced module (CIBM) and commonly used methods. While the green part (c) is a common way to tackle the class imbalance problem in terms of the number of samples for each category, our proposed CIBM takes into account the cosine similarity between samples within each category by adding the blue area (a) and the orange area (b).}
\label{fig:balance module}
\end{figure*}

In the Fig.\ref{fig:balance module}, while the green part (c) is a common way to consider the class imbalance problem in terms of the number of samples, our proposed CIBM takes into account the cosine similarity between samples of each category by adding feature extraction and similarity calculation.
We compare the relational differences between the traditional method for category balancing by means of sampling and the proposed CIBM.

Traditionally, as shown in Fig.\ref{fig:balance module}(c) part, we assume that the unbalanced dataset has in total $n$ individual categories, each of which has the number of samples ${x_1,x_2,... ,x_n}$, then in the training phase each category is assigned weights ${W_1, W_2,... ,W_n}$, so as to ensure that the number of samples for different categories is balanced throughout the process, and the weights are calculated as follows:

\begin{equation}
\label{eq:num_balance1}
    p_{i}=\frac{x_{i}}{\sum_{i=1}^{n} x_{n}}, 
\end{equation}

\begin{equation}
\label{eq:num_balance2}
    w_{i}=\frac{p_{i}^{-1}}{\sum_{i=1}^{n} p_{i}^{-1}}.
\end{equation}
where ${p_i}$ represents the number of samples in category ${i}$ as a proportion of the total number of samples, and ${w_i}$ is the normalized weight.

The design of CIBM includes the three modules in Fig.\ref{fig:balance module}(a), (b), and (c), taking into account information about different samples within each category as well. As shown in (a), we feed samples from each category into the decoder of existing pre-trained model for category information extraction, which can be described by the following expression:

\begin{equation}
\label{eq:num_balance3}
    f(i,j)=D (I(i, j)).
\end{equation} 

where ${D()}$ represents the decoding operation to extract category feature vector, ${I(i,j)}$ represents ${j^{th}}$ sample in ${i^{th}}$ category, and ${f(i,j)}$ denotes the features extracted from the corresponding image.

And in Fig.\ref{fig:balance module}(b) we calculate the Euclidean feature distance between any two samples within each category, and finally for each category a Euclidean distance matrix is generated, which is calculated as below:

\begin{equation}
\label{eq:num_balance4}
    \operatorname{\textit{dis}}(\textit{i}, \textit{j}, \textit{k})=\sqrt{\sum_{x=1}^{d_{} }\left(f(i, j)_{x} -f(i, k)_{x})^{2}\right.}
\end{equation}
where ${dis}(i, j, k)$ represents the value of the ${j^{th}}$ row ${k}$ columns in the distance matrix of the ${i^{th}}$ category and ${d}$ is the length of the category feature vector.

So the category distance weights ${S_i}$ and the final sampling weights can be obtained as follows:

\begin{equation}
\label{eq:num_balance51}
    S_{i}=\sum_{j=1}^{x_{i}} \sum_{k=1}^{x_{i}} \operatorname{\textit{dis}(\textit{i}, \textit{j}, \textit{k}),(\textit{j}\neq \textit{k})}, 
\end{equation}

\begin{equation}
\label{eq:num_balance52}
    W_{i}=\frac{S_{i} \cdot p_{i}^{-1}}{\sum_{i=1}^{n}\left(S_{i} \cdot p_{i}^{-1}\right)}.
\end{equation}

\subsubsection{\textbf{Loss Function}}
The cross-entropy loss with respect to the output features of the pretrained model is shown below:
\begin{equation}
    L_{con}=-l o g(\frac{e x p(z|c])}{\Sigma_{j=0}^{c-1}\:e x p(z|j))})=-z[c]+l o g(\sum_{j=0}^{C-1}\:e x p(z[j]))
\end{equation}

The cross-entropy loss with respect to the ground truth is denoted below:
\begin{equation}
    {L_{cls}}=-\sum_{i=0}^{C-1}y_{i}l o g(p_{i})=-l o g(p_{c})
\end{equation}
The final loss function is the weighted sum of the above loss functions:
\begin{equation}
    Loss = \alpha L_{con} + (1-\alpha) L_{cls}
\end{equation}
where $\alpha$ is in the middle of the interval [0, 1], and when $\alpha=0$, no contrastive loss is added, the loss function is simply ground-truth supervised, in line with the traditional approach. In our experiments, $\alpha$ is set as 0.7.

\section{Experiment}
\label{sec:exp}
\subsection{\textbf{Dataset}}
\label{sec:dataset}

\begin{figure*}[htbp]
  \centering
  \begin{minipage}[b]{0.45\linewidth}
    \centering
    \includegraphics[width=\linewidth]{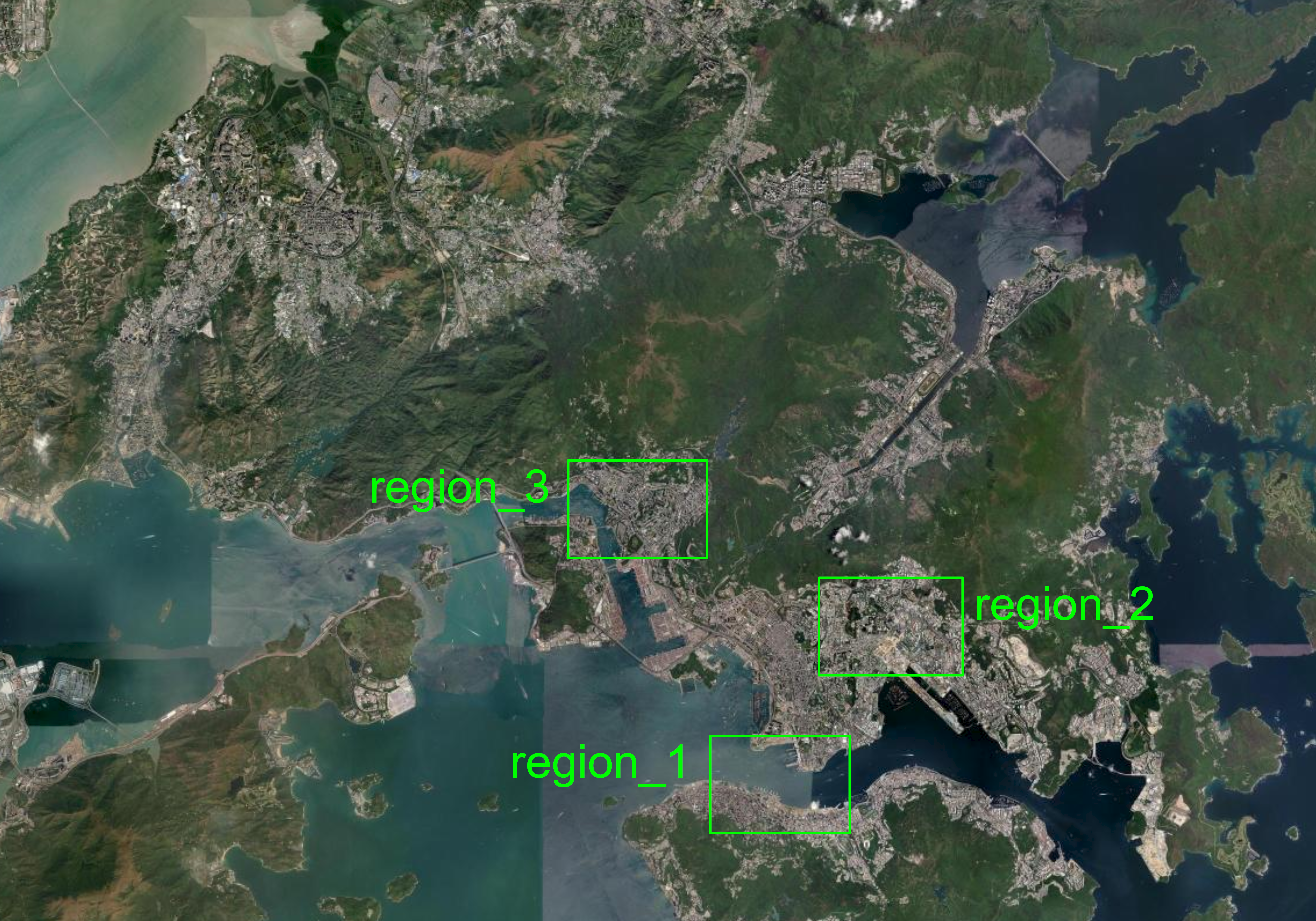}
  \end{minipage}%
  \hspace{0.03\linewidth} 
  \begin{minipage}[b]{0.45\linewidth}
    \centering
    \includegraphics[width=\linewidth]{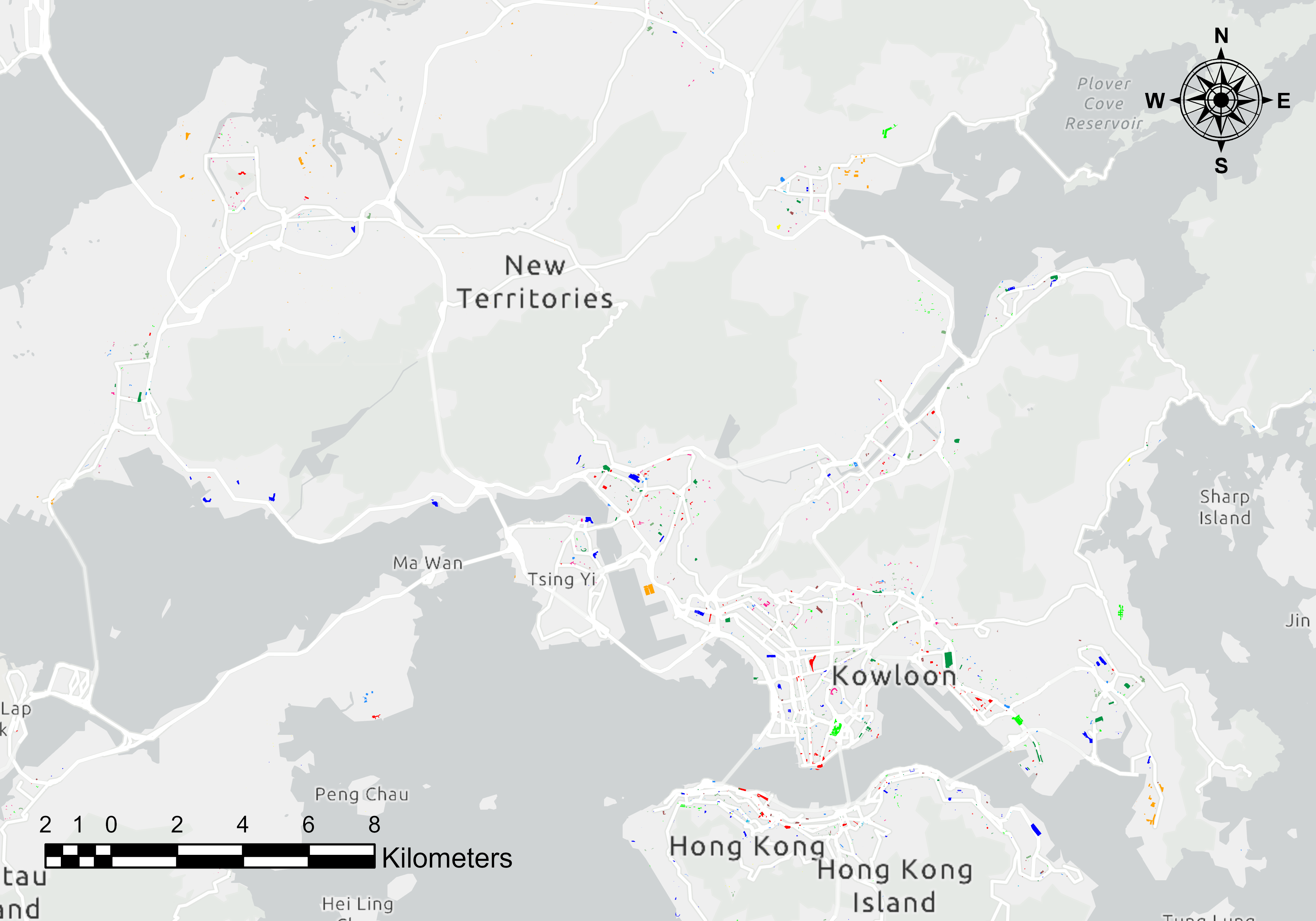}
  \end{minipage}%
  
\vspace{0.02\linewidth}

 \begin{minipage}[b]{0.6\linewidth}
    \centering
    \includegraphics[width=\linewidth]{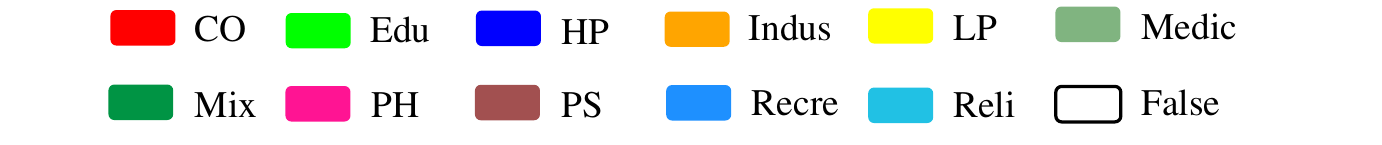}
  \end{minipage}
  
  \caption{Research region. }
  \label{fig:HK_images}
\end{figure*}

\begin{figure}[t]
\begin{center}
\includegraphics[width=0.45\textwidth]{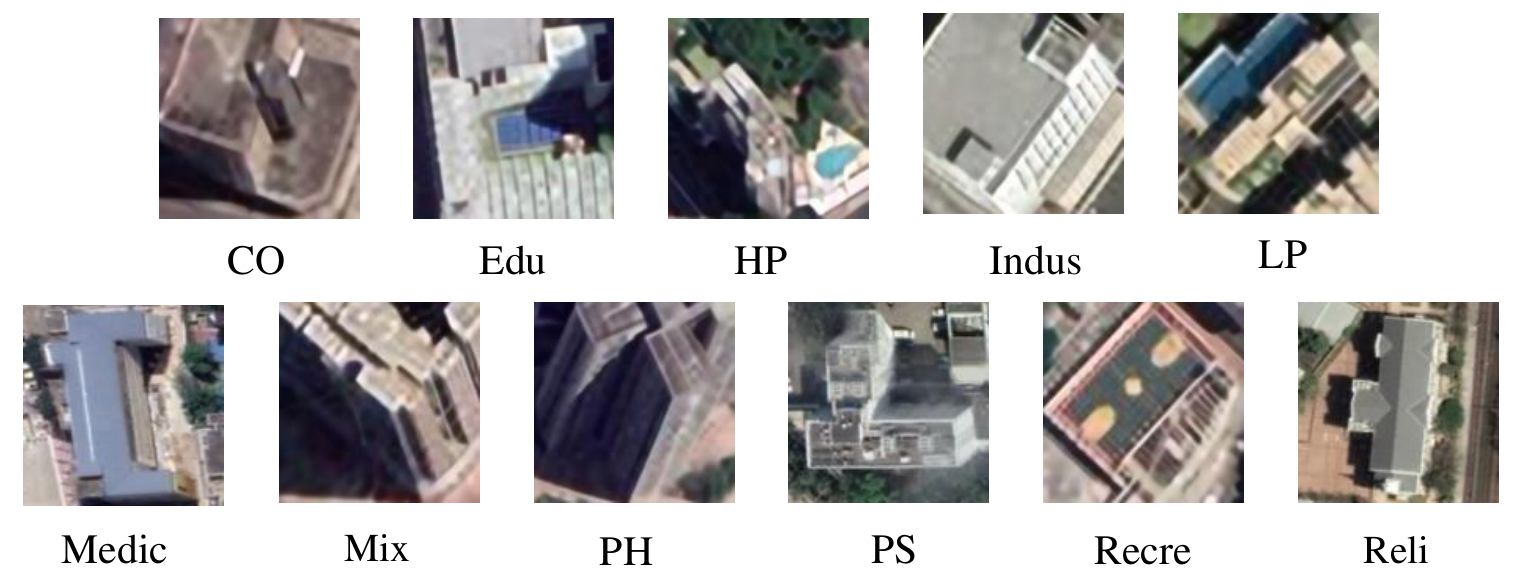}
\end{center}
\caption{Fine-grained building category samples.}
\label{fig:all_classes_images}
\end{figure}

\begin{table*}[h]
\caption{Fine-grained building categories.}
\label{tab:building categories}
\centering
\begin{tabular}{c c c}
\toprule
Category names & Abbreviations & Examples \\
\midrule
Commercial \& Office building & CO & offices, retail stores, and shopping centers \\
Educational institution & Edu & schools, colleges, universities, and other educational institutions \\
High-rise private housing & HP & hight residential buildings,condominiums \\
Industrial building & Indus & factories, warehouses, manufacturing plants, and distribution centers \\
Low-rise private housing & LP & low residential buildings \\
Medical building & Medic &  hospitals, clinics, medical offices, and healthcare facilities \\
Mixed-used building & Mix &  a combination of residential, commercial, and/or office spaces \\
Public rental housing & PH & public residential buildings \\
Public services building & PS & government offices, public libraries, post offices, and community centers \\
Recreation & Recre & amusement parks, and entertainment venues  \\
Religious facility & Reli &  churches, temples, mosques \\

\bottomrule
\end{tabular}
\end{table*}

\vspace*{-0.08cm}
\begin{table*}[b]
 \caption{Comparison of building classification results obtained from different methods, including following metrics \textit{Top-1 Acc, Top-5 Acc, Mean Precision, Mean Recall, Mean F1 Score}, $\uparrow$ means higher is better.}
 \label{tab:classification results}
    \centering
    \small
\scalebox{.95}
    {    
    \begin{tabular}{l@{\hspace{.2cm}}c@{\hspace{.2cm}}c@{\hspace{.2cm}}c@{\hspace{.2cm}}c@{\hspace{.2cm}}c@{\hspace{.2cm}}c}
    \hline
    \textbf{Metric} &  Top-1 Acc $\uparrow$ &  Top-5 Acc $\uparrow$ &    Mean Precision $\uparrow$ & Mean Recall $\uparrow$ &  Mean F1 Score $\uparrow$  &  Model size $\downarrow$\\
    \hline
    \text{MVit} & 44.64 & 87.95 & 44.34 & 44.64 & 44.03 & 609.4Mb\\
    \text{EfficientFormer} & 38.68 & 85.32 & 38.05 & 38.68 & 37.75 & 137.3Mb\\
    \text{EfficientNet-b0} & 49.91 & 90.68 & 50.26 & 49.91 & 48.92 & 32.7Mb\\
    \text{EfficientNet-b5} & 51.05 & 90.91 & 50.36 & 51.05 & 50.03 & 228.2Mb\\
    \text{ShuffleNetV2(baseline)} & 53.05 & 90.95 & 52.68 & 53.64 & 53.46 & \textbf{11.1}Mb\\
    \hline
    \textbf{Our method} & \textbf{60.45}$^{(+13.9\%)}$ & \textbf{93.50}$^{(+2.8\%)}$ & \textbf{60.57}$^{(+15.0\%)}$ & \textbf{60.45}$^{(+12.6\%)}$ & \textbf{60.47}$^{(+13.1\%)}$ & \textbf{11.1}Mb\\
    \hline
    \end{tabular}
    }
    \vspace*{0.1cm}
    \vspace*{-0.1cm}
\end{table*}

Buildings have a wide variety of functions and are often related to factors such as the level of local economic development and religion. The main area of study in this paper is the Hong Kong SAR, as shown in the Fig.\ref{fig:HK_images}. We classify the buildings in Hong Kong into 11 main categories according to their functions (Table \ref{tab:building categories}: commercial and office building, educational institution, high-rise private housing, industrial building, low-rise private housing, medical building, mixed-used building, public rental housing, public service/government building, recreation, religious facility), with each category containing 1,000 images and there are 11,000 images in total. We acquired the data in a similar way as previous research\citep{TONG2020111322}, intercepting the images as 32$\times$32 chunks with a spatial resolution of 4.78m from Google Earth. And Fig.\ref{fig:all_classes_images} shows the given 11 category samples after 4$\times$ super-resolution. 

We use building function as reference and divide all the data into 5 equal parts for each category, with each part of 2200 samples, the model training step takes four parts of each category as training data and the rest one part as test data, the data ratio is 4:1. This enables to make the full use of each sample and also improves the generalization of the network.

Test images are satellite overhead images intercepted from Google Earth based on geo-referenced coordinates of building instances, where coordinates are obtained from the Hong Kong Government's public GeoData Store \footnote{\url{https://geodata.gov.hk/gs/}}. Due to the inconsistent size and shape of buildings and the small number of pixels occupied in the satellite images, the original intercepted images are of low resolution. The example image is shown in the Fig.~\ref{fig:all_classes_images}. Top-view observation of buildings in low resolution satellite images is a big challenge for feature extraction of the fine classification network, as inconspicuous and similar features can lead to significant performance loss or even failure.

\subsection{\textbf{Training Details}}
\label{sec:training}
The method proposed in this paper trains two models, the first one is DDPM-based super-resolution model for satellite images and the other one is classification network that jointly learns the building function and age. Both models are trained with two 2080Ti GPUs.

\subsubsection{\textbf{Super-Resolution}}
\label{sed:DDPM training}
A total of 14,292 training images and 6,357 validation images were obtained from satellite images of Hong Kong, all at a resolution of about 4.78m and size of 32$\times$32, and a HR image  with resolution of about 1.195m and size of 128$\times$128 pixels which acts as super-resolution ground truth. In the experiments, we set the number of time step to 2000 in both training and validation period, the start and end liner parameters were set as 1$\times 10^{-6}$ and 1$\times 10^{-2}$ separately.


\subsubsection{\textbf{Building Classification}}
\label{classification training}

We use ShuffleNetV2 \citep{EfficientNet2019} as our backbone, which is pre-trained on ImageNet \citep{imagenet2009}. We use the Adam optimizer \citep{Kingma2014AdamAM} without weight decay and decrease the learning rate from 1.5$\times 10^{-2}$ to 1$\times 10^{-5}$ by the step down scheduler. To avoid over-fitting, the images are augmented by horizontal flipping and random crop. Our models are trained for 50 epoches.

\subsection{\textbf{Evaluation Metrics}}

The metrics that measure the performance of the classification method in this study are Top-1 Accuracy, Top-5 Accuracy,  Mean Precision,  Mean Recall and Mean F1 Score. They are calculated as follows:\\

 Top-1 Accuracy $=$ $\dfrac{n_{1}}{N_{t}}$,   
 Top-5 Accuracy $=$ $\dfrac{n_{5}}{N_{t}}$\\ 

 ($n_{1}$ represents Top-1 True Positive number, $n_{5}$ represents Top-5 True Positive number, $N_{t}$ represents test number)\\

 Mean Precision $=$ $\dfrac{\text{True Positive}}{\text{True Positive}+\text{False Positive}}$\\
 
 Mean Recall $=$ $\dfrac{\text{True Positive}}{\text{True Positive}+\text{False Negative}}$\\
 
 Mean F1 Score $=$ $2\cdot\dfrac{\text{Precision}\cdot \text{Recall}}{\text{Precision}+\text{Recall}}$

\section{Results and Discussion}
\label{sec:exp results}

\subsection{\textbf{Super-Resolution}}

\begin{figure}[t]
\begin{center}
\includegraphics[width=0.5\textwidth]{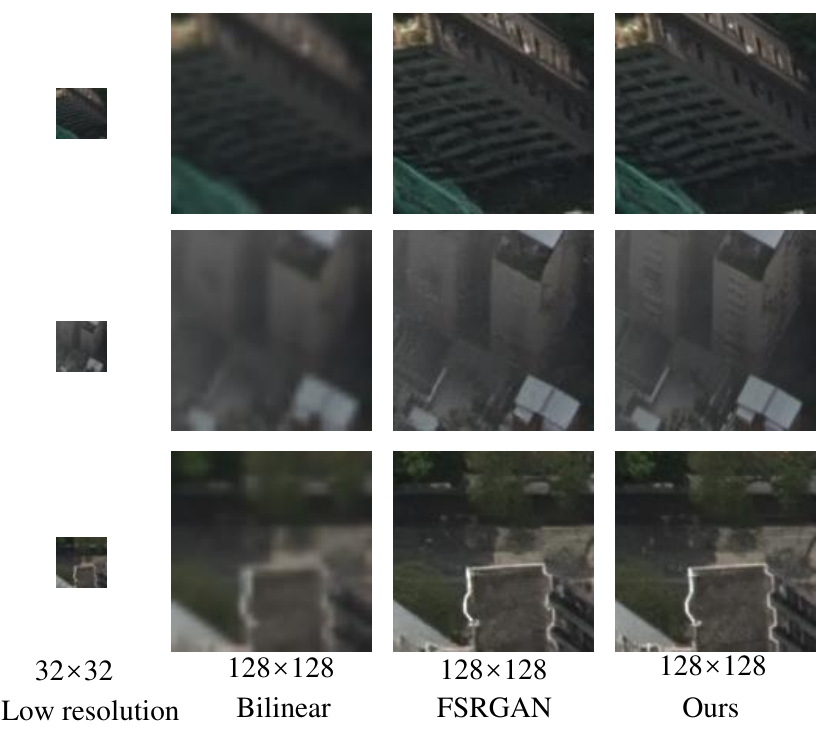}
\end{center}
\caption{Comparison of super-resolution images using different methods.}
\label{fig:SR_results}
\end{figure}

The results are shown in Fig.\ref{fig:SR_results}, the intercepted original satellite image is displayed in the left column, which is too small to be directly classified by the network. The right three columns show the results after 4-fold super-resolution using the Bilinear, FSRGAN, and the method proposed in this paper, respectively. Although the image sizes have been upgraded to 128×128, there are significant differences in the retained architectural structure and associated feature information. It is evident that, following the implementation of the method proposed in this paper, the building roof outlines and side details are more distinctive. As shown in Tab.\ref{tab:psnr_ssim_images}, our DDPM based method perform better than other super-resolution methods on PSNR, SSIM and Consistency. Additionally, the super-resolution effect surpasses that of the first two methods. This finding holds crucial significance for the Phase2 classification and will be elaborated upon in the ablation experiment section.

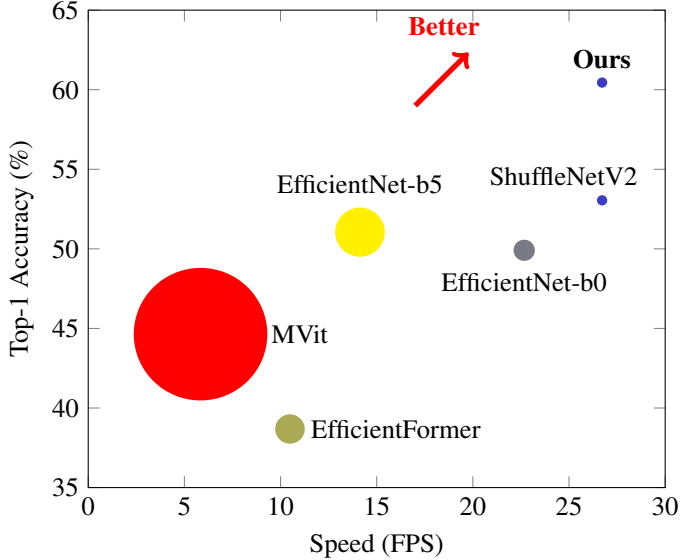
\begin{figure}
  \centering
  \begin{tikzpicture}
    \begin{axis}[
      width=0.5\textwidth,
      xlabel={Speed (FPS)},
      ylabel={Top-1 Accuracy (\%)},
      xmin=0, xmax=30,
      ymin=35, ymax=65,
      ]
      \addplot[
        scatter,
        only marks,
        scatter src=explicit,
        scatter/use mapped color={draw opacity=0,fill=mapped color},
        scatter/@pre marker code/.append style={
          /tikz/mark size=\pgfplotspointmetatransformed/40
        },
        nodes near coords align={below},
        nodes near coords style={
          font=\footnotesize,
        },
      ] table[x=Speed, y=Accuracy, meta=Parameters] {
        Speed Accuracy Parameters
        5.84 44.64 609.4
        10.49 38.68 137.3
        22.67 49.91 100
        14.13 51.05 228.2
        26.72 53.05 50
        26.72 60.45 50
        26.72 60.45 3.11
      };
      
    \node at (axis cs:11,44.64) {MVit};
    \node at (axis cs:16,38.68) {EfficientFormer};
    \node at (axis cs:22.67,47.91) {EfficientNet-b0};
    \node at (axis cs:14.13,54.05) {EfficientNet-b5};
    \node at (axis cs:24.72,54.55) {ShuffleNetV2};
    \node at (axis cs:26.72,61.95) {\textbf{Ours}};
    
    \node at (axis cs:18.5,64) {\textcolor{red}{\textbf{Better}}};
    \draw[->, line width=2pt, red] (axis cs:17,59) -- ++(0.7cm,0.7cm);
    
    \end{axis}
  \end{tikzpicture}

  \caption{Top-1 accuracy, inference speed and model size on the test set. The size of the circle denotes the model size, the smaller it is the fewer the model parameters. The closer the model to top right corner represents higher accuracy and faster inference. As the red arrow shows, it can be perceived as toward better performance.}
\label{fig:visualisation of comparative results}
\end{figure}

\vspace*{-0.08cm}
\begin{table}[h]
 \caption{PSNR \& SSIM on  32$\times$32 $\rightarrow$ 128$\times$128 satellite image super-resolution.
    Consistency measures MSE ($\times 10^{-5}$) between the low-resolution inputs and the down-sampled super-resolution outputs, $\uparrow$ means higher is better and $\downarrow$ means lower is better}
\label{tab:psnr_ssim_images}
    \centering
    \small
\scalebox{.95}{    
    \begin{tabular}
    {l@{\hspace{.2cm}}c@{\hspace{.2cm}}c@{\hspace{.2cm}}c@{\hspace{.2cm}}c}
    \hline
    \bfseries Metric  & \bfseries FSRGAN & \bfseries Regression & \bfseries Our method \\
    \hline
    \textbf{PSNR} $\uparrow$ & 23.01 & 23.04 & {\bf 23.96}\\
    \textbf{SSIM} $\uparrow$ & 0.62  & 0.65 & {\bf 0.69}\\
    \hline
    \textbf{Consistency} $\downarrow$ & 33.8 & 2.71 & {\bf 2.68} \\
    \hline
    \end{tabular}
    }
    \vspace*{0.1cm}
    \vspace*{-0.1cm}
\end{table}

\subsection{\textbf{Building Function Classification}}

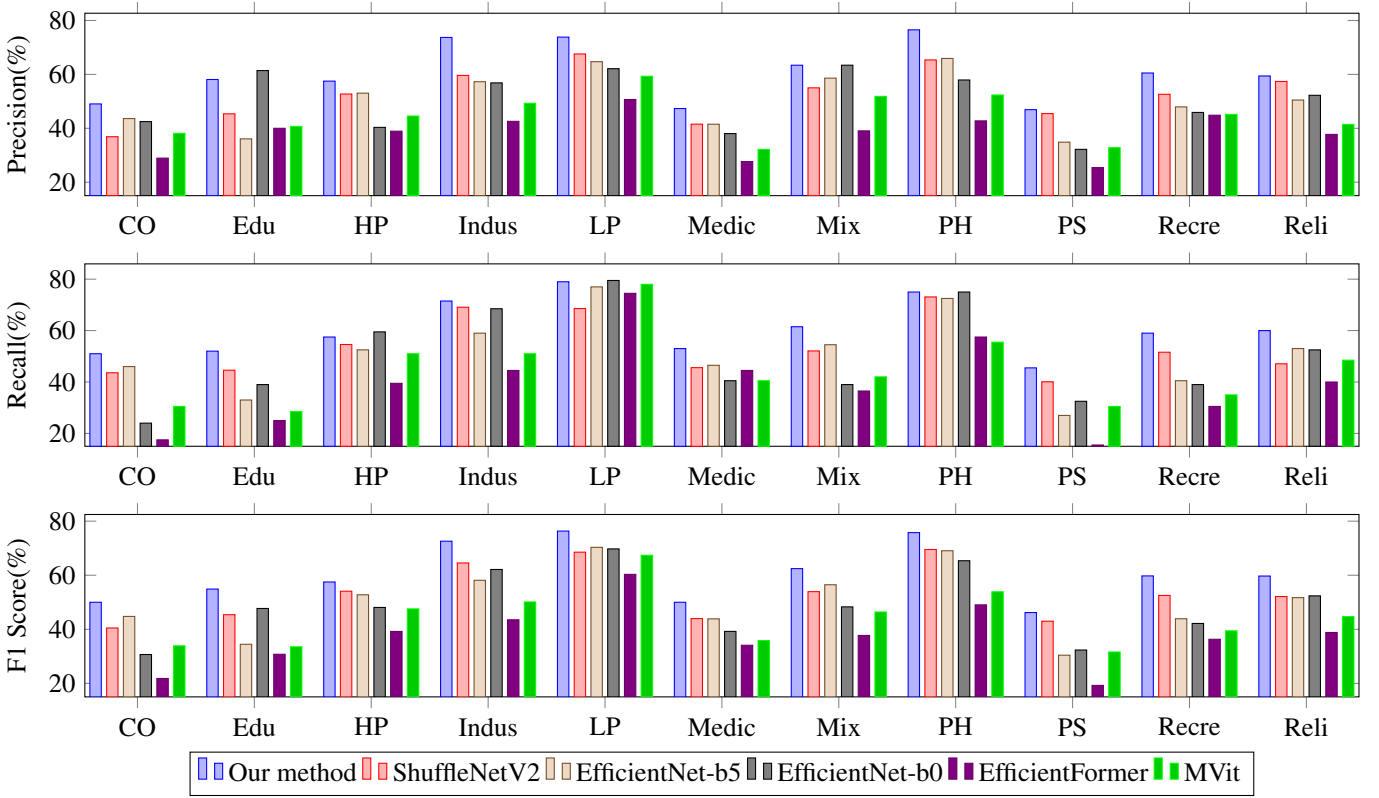
\begin{figure*}[b]
    \centering

\begin{minipage}{\textwidth}
        \centering
        \begin{tikzpicture}
            \begin{axis}[
                ybar,
                bar width=0.15cm,
                width=\textwidth,
                height=4cm,
                ylabel={Precision($\%$)},
                ymin=15,
                symbolic x coords={CO, Edu, HP, Indus, LP, Medic, Mix, PH, PS, Recre, Reli},
                xtick=data,
                legend style={at={(0.5,-0.2)}, anchor=north, legend columns=-1},
                enlarge x limits=0.045,
            ]
           \addplot coordinates {(CO, 49.04) (Edu, 58.1) (HP, 57.5) (Indus, 73.71) (LP, 73.83) (Medic, 47.32) (Mix, 63.4) (PH, 76.53) (PS, 46.91) (Recre, 60.51) (Reli, 59.41)};
        
            \addplot coordinates {(CO, 36.87) (Edu, 45.36) (HP, 52.71) (Indus, 59.62) (LP, 67.58) (Medic, 41.53) (Mix, 55.01) (PH, 65.36) (PS, 45.44) (Recre, 52.61) (Reli, 57.38)};

            \addplot coordinates {(CO, 43.6) (Edu, 36.07) (HP, 53.03) (Indus, 57.28) (LP, 64.71) (Medic, 41.52) (Mix, 58.6) (PH, 65.91) (PS, 34.84) (Recre, 47.93) (Reli, 50.48)};

            \addplot coordinates {(CO, 42.48) (Edu, 61.42) (HP, 40.34) (Indus, 56.85) (LP, 62.11) (Medic, 38.03) (Mix, 63.41) (PH, 57.92) (PS, 32.18) (Recre, 45.88) (Reli, 52.24)};

             \addplot coordinates {(CO, 28.93) (Edu, 40) (HP, 38.92) (Indus, 42.58) (LP, 50.68) (Medic, 27.64) (Mix, 39.04) (PH, 42.75) (PS, 25.41) (Recre, 44.85) (Reli, 37.74)};

            \addplot coordinates {(CO, 38.12) (Edu, 40.71) (HP, 44.54) (Indus, 49.28) (LP, 59.32) (Medic, 32.14) (Mix, 51.85) (PH, 52.36) (PS, 32.8) (Recre, 45.16) (Reli, 41.45)};

            \end{axis}
        \end{tikzpicture}
    \end{minipage}
    
    \vspace{0.08cm}

\begin{minipage}{\textwidth}
        \centering
        \begin{tikzpicture}
            \begin{axis}[
                ybar,
                bar width=0.15cm,
                width=\textwidth,
                height=4cm,
                ylabel={Recall($\%$)},
                ymin=15,
                symbolic x coords={CO, Edu, HP, Indus, LP, Medic, Mix, PH, PS, Recre, Reli},
                xtick=data,
                legend style={at={(0.5,-0.2)}, anchor=north, legend columns=-1},
                enlarge x limits=0.045,
            ]
           \addplot coordinates {(CO, 51) (Edu, 52) (HP, 57.5) (Indus, 71.5) (LP, 79) (Medic, 53) (Mix, 61.5) (PH, 75) (PS, 45.5) (Recre, 59) (Reli, 60)};

            \addplot coordinates {(CO, 43.59) (Edu, 44.59) (HP, 54.59) (Indus, 69.09) (LP, 68.59) (Medic, 45.59) (Mix, 52.09) (PH, 73.09) (PS, 40.09) (Recre, 51.59) (Reli, 47.09)};

            \addplot coordinates {(CO, 46) (Edu, 33) (HP, 52.5) (Indus, 59) (LP, 77) (Medic, 46.5) (Mix, 54.5) (PH, 72.5) (PS, 27) (Recre, 40.5) (Reli, 53)};

            \addplot coordinates {(CO, 24) (Edu, 39) (HP, 59.5) (Indus, 68.5) (LP, 79.5) (Medic, 40.5) (Mix, 39) (PH, 75) (PS, 32.5) (Recre, 39) (Reli, 52.5)};

             \addplot coordinates {(CO, 17.5) (Edu, 25) (HP, 39.5) (Indus, 44.5) (LP, 74.5) (Medic, 44.5) (Mix, 36.5) (PH, 57.5) (PS, 15.5) (Recre, 30.5) (Reli, 40)};

            \addplot coordinates {(CO, 30.5) (Edu, 28.5) (HP, 51) (Indus, 51) (LP, 78) (Medic, 40.5) (Mix, 42) (PH, 55.5) (PS, 30.5) (Recre, 35) (Reli, 48.5)};

            \end{axis}
        \end{tikzpicture}
    \end{minipage}
    
    \vspace{0.08cm}
 
\begin{minipage}{\textwidth}
        \centering
        \begin{tikzpicture}
            \begin{axis}[
                ybar,
                bar width=0.15cm,
                width=\textwidth,
                height=4cm,
                ylabel={F1 Score($\%$)},
                ymin=15,
                symbolic x coords={CO, Edu, HP, Indus, LP, Medic, Mix, PH, PS, Recre, Reli},
                xtick=data,
                legend style={at={(0.5,-0.3)}, anchor=north, legend columns=-1},
                enlarge x limits=0.045,
            ]
            \addplot coordinates {(CO, 50) (Edu, 54.88) (HP, 57.5) (Indus, 72.59) (LP, 76.33) (Medic, 50) (Mix, 62.44) (PH, 75.76) (PS, 46.19) (Recre, 59.75) (Reli, 59.7)};

            \addplot coordinates {(CO, 40.49) (Edu, 45.4) (HP, 54.09) (Indus, 64.54) (LP, 68.53) (Medic, 43.96) (Mix, 53.93) (PH, 69.52) (PS, 43) (Recre, 52.53) (Reli, 52.12)};

            \addplot coordinates {(CO, 44.77) (Edu, 34.46) (HP, 52.76) (Indus, 58.13) (LP, 70.32) (Medic, 43.87) (Mix, 56.48) (PH, 69.05) (PS, 30.42) (Recre, 43.9) (Reli, 51.71)};

            \addplot coordinates {(CO, 30.67) (Edu, 47.71) (HP, 48.08) (Indus, 62.13) (LP, 69.74) (Medic, 39.23) (Mix, 48.3) (PH, 65.36) (PS, 32.34) (Recre, 42.16) (Reli, 52.37)};

            \addplot coordinates {(CO, 21.81) (Edu, 30.77) (HP, 39.21) (Indus, 43.52) (LP, 60.32) (Medic, 34.1) (Mix, 37.73) (PH, 49.04) (PS, 19.25) (Recre, 36.31) (Reli, 38.83)};

            \addplot coordinates {(CO, 33.89) (Edu, 33.53) (HP, 47.55) (Indus, 50.12) (LP, 67.39) (Medic, 35.84) (Mix, 46.41) (PH, 53.88) (PS, 31.61) (Recre, 39.44) (Reli, 44.7)};
        
            \legend{Our method, ShuffleNetV2, EfficientNet-b5, EfficientNet-b0, EfficientFormer, MVit}
            \end{axis}
        \end{tikzpicture}
    \end{minipage}
    
    \caption{Fine building classification results of our model and various SOTA models. The top, middle and bottom rows represent the results of the Precision, Recall and F1 Score indicators respectively.}
    \label{fig:categories matrics comparison}
\end{figure*}
\vspace*{-0.08cm}
\begin{table}[h]
 \caption{Confusion matrix of classification results.}
 \label{tab:confusion matrix}
    \centering
    \small
\scalebox{.95}{    
    \begin{tabular}
    {l@{\hspace{.15cm}}c@{\hspace{.15cm}}c@{\hspace{.15cm}}c@{\hspace{.15cm}}c@{\hspace{.15cm}}c@{\hspace{.15cm}}c@{\hspace{.15cm}}c@{\hspace{.15cm}}c@{\hspace{.15cm}}c@{\hspace{.15cm}}c@{\hspace{.15cm}}c}
    \hline
    \textbf {Class} & CO & Edu & HP & Indus & LP & Medic & Mix & PH & PS & Recre & Reli\\
    
    \hline
    CO & \textbf{102} & 15 & 13 & 3 & 6 & 17 & 14 & 5 & 13 & 10 & 2\\
    Edu & 16 & \textbf{104} & 8 & 8 & 7 & 19 & 5 & 5 & 11 & 12 & 5\\
    HP & 17 & 4 & \textbf{115} & 1 & 7 & 5 & 24 & 18 & 4 & 3 & 2\\
    Indus & 7 & 1 & 0 & \textbf{143} & 12 & 3 & 0 & 0 & 14 & 13 & 7\\
    LP & 0 & 6 & 10 & 10 & \textbf{158} & 4 & 3 & 2 & 1 & 2 & 4\\
    Medic & 7 & 13 & 2 & 6 & 0 & \textbf{106} & 9 & 3 & 26 & 8 & 20\\
    Mix & 15 & 4 & 29 & 1 & 7 & 7 & \textbf{123} & 7 & 2 & 3 & 2\\
    PH & 18 & 6 & 14 & 0 & 7 & 1 & 0 & \textbf{150} & 3 & 0 & 1\\
    PS & 16 & 9 & 4 & 3 & 3 & 34 & 3 & 4 & \textbf{91} & 13 & 20\\
    Recre & 7 & 11 & 1 & 11 & 4 & 10 & 3 & 1 & 15 & \textbf{118} & 19\\
    Reli & 3 & 6 & 4 & 8 & 3 & 18 & 10 & 1 & 14 & 13 & \textbf{120}\\
    \hline
    \end{tabular}
    }
    \vspace*{0.1cm}
    \vspace*{-0.1cm}
\end{table}

\begin{figure*}[h]
  \centering
   \begin{minipage}[b]{1.0\linewidth}
    \centering
    \includegraphics[width=\linewidth]{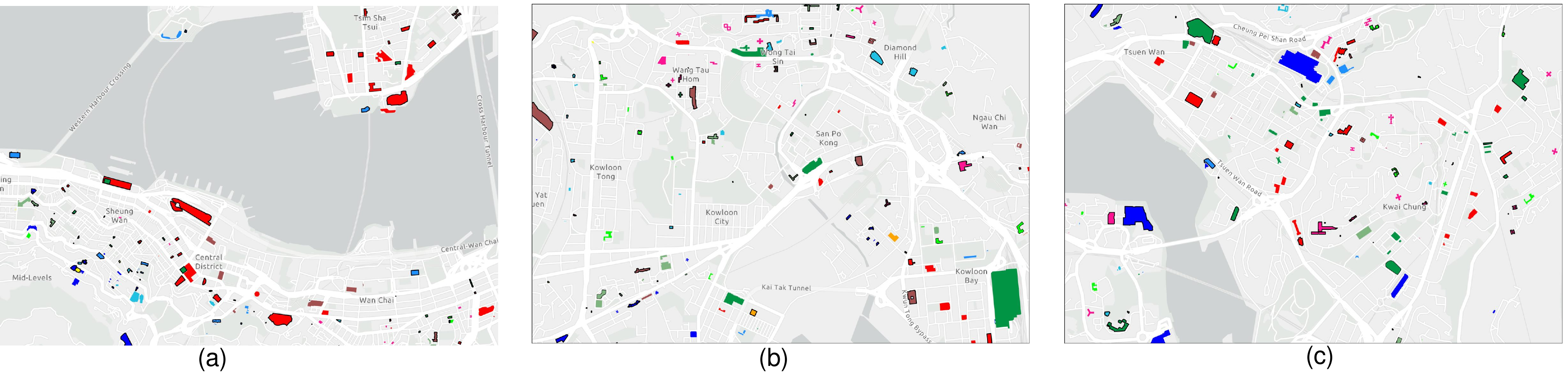}
  \end{minipage}
 \begin{minipage}[b]{0.6\linewidth}
    \centering
    \includegraphics[width=\linewidth]{color_legend.pdf}
  \end{minipage}
  \caption{Visualisation of building category classification results.}
  \label{fig:test_images}
\end{figure*}

\vspace*{-0.08cm}
\begin{table*}[b]
 \caption{Comparison of results of different super-resolution methods.}
 \label{tab:abla ddpm}
    \centering
    \small
\scalebox{.95}{    
    \begin{tabular}
    {l@{\hspace{.2cm}}c@{\hspace{.2cm}}c@{\hspace{.2cm}}c@{\hspace{.2cm}}c@{\hspace{.2cm}}c}
    \hline
    Metric  &  Top-1 Acc $\uparrow$ &   Top-5 Acc $\uparrow$ &    Mean Precision $\uparrow$ & Mean Recall $\uparrow$ &  Mean F1 Score $\uparrow$ \\
    \hline
        \text{Bicubic+CIBM+CS} & 50.25 & 89.67 & 51.91 & 51.89 & 50.06\\
        \text{Regression+CIBM+CS}  & 53.51 & 90.92 & 54.13 & 53.28 & 54.27\\
        \text{FSRGAN+CIBM+CS} & 56.77 & 91.94 & 55.81 & 56.25 & 56.12\\
    \hline
    \textbf{DDPM+CIBM(Ours)} & 55.24 & 91.12 &54.87 & 55.08 & 54.96\\
    \textbf{DDPM+CIBM+CS(Ours)}  & \textbf{60.45 }&\textbf{93.50} & \textbf{60.57} & \textbf{60.45} & \textbf{60.47}\\
    \hline
    \end{tabular}
    }
    \vspace*{0.1cm}
    \vspace*{-0.1cm}
\end{table*}

\begin{figure*}[!htbp]
\begin{center}
\includegraphics[width=0.95\textwidth]{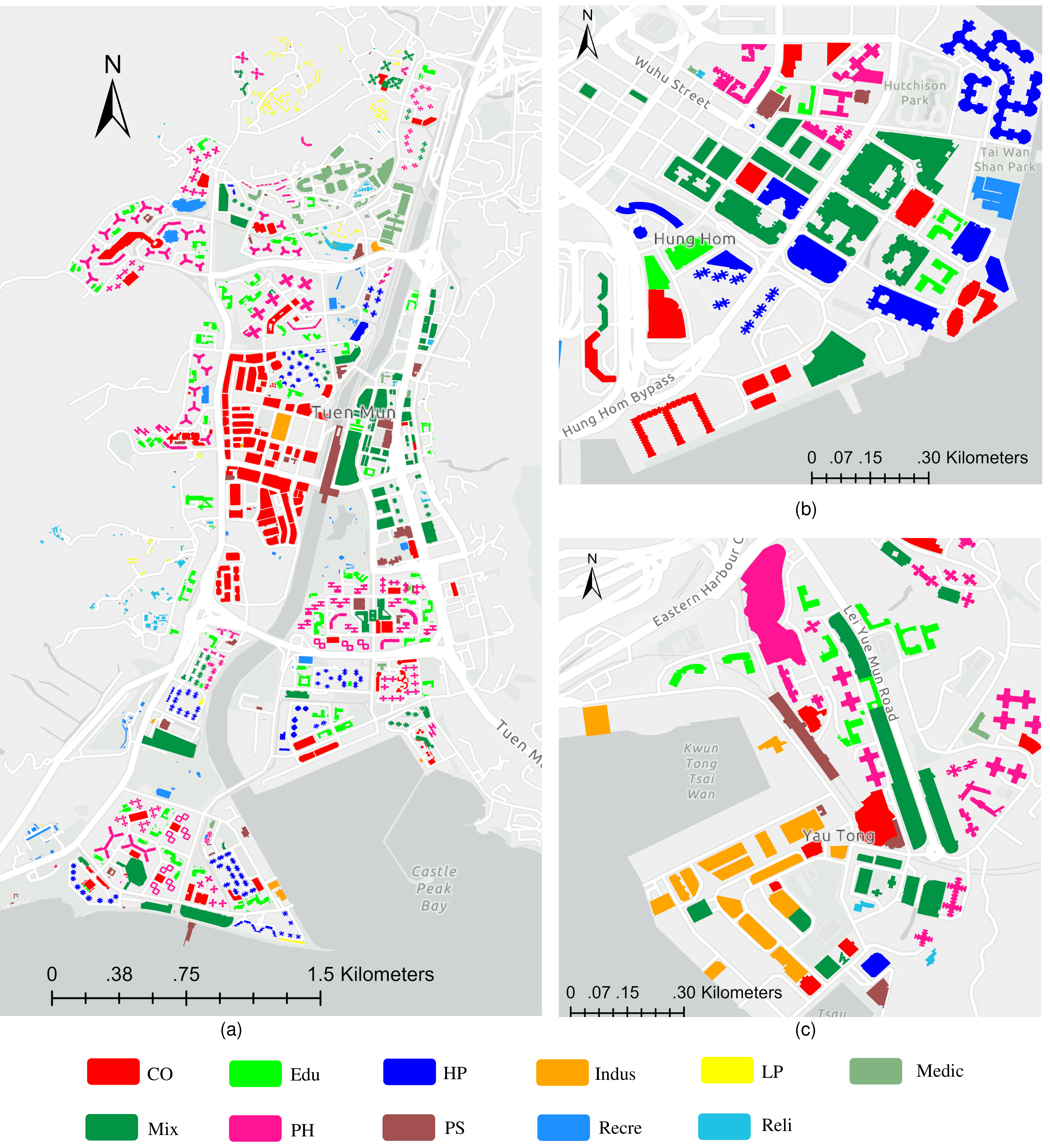}
\end{center}
\caption{Open test results}
\label{fig:open test results}
\end{figure*}

In order to verify the validity of our method, we have conducted experiments on our data using MVit, EfficientFormer, EfficientNet and ShuffleNetV2, which are currently state-of-art methods for image classification and compared in details to our method. The classification results are shown in Table \ref{tab:classification results}, for the sake of fairness, the results in the table are generated based on the super-resolved images of the DDPM method, while the postal validity and necessity of the DDPM method are discussed in the ablation experiment section. As can be seen from Table \ref{tab:classification results}, MVit and EfficientFormer, based on the transformer approach, do not perform as well as EfficientNet and ShuffleNetV2, based on the CNN approach, in terms of Acc, Precision, Recall and F1 Score results on this dataset, which is consistent with the characteristics of these two deep network families. Transformer-structured networks tend to perform well in scenarios with high data quality and large data volumes, such as ImageNet and COCO, and the effectiveness decreases when the data volume is small, as for the case of our dataset. As shown in Fig.\ref{fig:visualisation of comparative results}, note that CNN-based networks have a smaller Model Size, which means that they have fewer parameters to learn and can be trained to produce good parametric models with less data while remain lightweight. The Model Size of our model is only 11.1Mb, which is one 54$^{th}$ of MVit, one 12$^{th}$ of EfficientFormer, and one third of EfficientNet's lightest b0. Our model size is the same as the baseline (ShuffleNetV2), but with a large improvement in performance, by up to 14.8\%, 15.7\%, 12.6\% and 16.2\% in terms of Top-1 Acc, Mean Precision, Mean Recall and Mean F1 Score, respectively. In Fig.\ref{fig:categories matrics comparison}, we show the classification Top-1 Acc for each category from the different competing methods, and it can be seen that our method performs significantly better than the other methods in terms of accuracy for each sub-category, although baseline methods even perform well in some of categories. It is also important to note that due to considering the category information and balancing constraint by our method, the results for the 11 categories are relatively less volatile and the model is more stable.


The details of results for each category are shown in the confusion matrix of Table \ref{tab:confusion matrix}. We can see that the classification results varies for different building categories, with public rental house(PH) and low-rise private building(LP) showing significantly better results than the other categories. Buildings contained within the categories that work well have similar features, and are more conducive to learning a model to discriminate between their features. The poorly performing building categories, for example, the PS contains several subcategory, such as government offices, public libraries, post offices, and community centres, etc. Although they belong to the same category of public service, the buildings in satellite images exhibit heterogeneous features, which is not conducive for the learnable network to discriminate their shared features.

Our model's input data requirements are lower than those of Street View image-based methods in terms of both spatial resolution and image size. The spatial resolution of satellite images used in this study is 4.78m, which is much lower than that of a typical Street View image. Additionally, the image size is 32$\times$32, as opposed to the Street View image size of 128$\times$128. Despite this, we have overcome this challenging problem and achieved better fine classification results than the Street View images \citep{ZHANG2023153}.


Three representative test areas of dense urban building clusters are depicted, demonstrating that the overall classification accuracy remains high, even though our method misclassified some samples (blocks with black outlines in Fig.\ref{fig:test_images}). Meanwhile, it is apparent that some of large buildings are misclassified while some of small ones are correctly classified. A possible explanation for this phenomenon is the relief displacement due to the building location relative to the image projection centre. When large buildings are located closer to the projection centre, the captured image is confined to the top-view of buildings with limited features, whereas when some of the small buildings are located farther away from the image projection centre, in addition to the roof features being preserved in the image, the side wall features are also preserved to a certain extent, which could be  useful for classification. This is because our model has been trained with the ability to handle this situation. Thus, the above perceptions seem counter-intuitive.


We conducted tests with application to the three test areas as demonstrated in Fig.\ref{fig:test_images} and Fig.\ref{fig:HK_images}, all of which exhibit high building density, complex building types, and random spatial distribution, thus posed as a challenging task. Our method's outcomes are illustrated in Fig.\ref{fig:open test results}, depicting the fine classification of building category utilizing solely low-resolution satellite images, without any omissions. 

\subsection{\textbf{Discussion}}
\label{sec:res}
%

\vspace*{-0.08cm}
\begin{table*}[h]
 \caption{Comparison of building classification datasets.}
 \label{tab:data}
    \centering
    \small
\scalebox{.95}
    {    
    \begin{tabular}{l@{\hspace{.2cm}}c@{\hspace{.2cm}}c@{\hspace{.2cm}}c@{\hspace{.2cm}}c@{\hspace{.2cm}}c@{\hspace{.2cm}}c@{\hspace{.2cm}}c}
    \hline
    \textbf{Data} & Source & Modality & View & Resolution & Size & Classed nums & Dense labelling\\
    \hline
    \text{UBC v1 \citep{HuangXingliang9857458}} & SuperView, GaoFen-2     & RGB & satellite view & 0.5-0.8m & 600 & 5  & Need\\
    \text{UBC v2 \citep{HuangXingliang10237272}}& SuperView, GaoFen-2/3   & RGB, SAR & satellite view & 0.5-0.8m & 512 & 12 & Need \\
    \hline
    \text{Ours} & Google Earth & RGB & satellite view & 4.78m & 32  & 11 & No need\\
    \hline
    \end{tabular}
    }
    \vspace*{0.1cm}
    \vspace*{-0.1cm}
\end{table*}

Due to various factors in research work, our proposed method and network produced promising results for the fine-grained building classification using overhead images, with a high improvement over the baseline. However, the absolute values of the classification results, e.g. Top-1 Acc (60.45\%),  are still relatively low compared to those of other conventional classification tasks, leaving quite room for improvement. Some ideas that could be further investigated: for example, a joint multimodal training framework can be built to improve the results by combining the overhead and street-view images or text information. The imbalance in the Confusion matrix is a good indicator of which features between categories are easily misidentified by the network. For the classification task with few similar categories, it may be inspired by the confusion matrix to design components to enhance the feature distinctiveness between similar categories and improve the network performance. In comparison, the alternative method necessitates high-resolution street-view images and still experiences omissions \citep{ZHANG2023153}. As shown in Tab.\ref{tab:data}, we compare the data requirements of implementing the latest methods for fine classification of urban buildings from satellite images\citep{HuangXingliang9857458,HuangXingliang10237272}, and it can be seen that our method requires images of lower spatial resolution. And the other ones are implemented through semantic segmentation, which requires time-consuming pixel-wise dense labels, whereas our method is exempted from this hassle.

\section{Ablation Study}
\label{sec:ablation study}
\begin{figure*}[b]
    \centering
    \begin{minipage}{0.48\textwidth}
        \centering
        \begin{tikzpicture}
            \begin{axis}[
                ybar,
                bar width=0.2cm,
                width=0.9\textwidth,
                height=5cm,
                ylabel={Top-1 Acc($\%$)},
                ymin=40,
                symbolic x coords={EfficientNet-b0, EfficientNet-b5, ShuffleNet-V2},
                xtick=data,
                legend style={at={(0.5,-0.2)}, anchor=north, legend columns=-1},
                enlarge x limits=0.15,
            ]
            \addplot [fill=gray] coordinates {(EfficientNet-b0, 49.91) (EfficientNet-b5, 51.05) (ShuffleNet-V2, 52.64)};
            \addplot [fill=orange] coordinates {(EfficientNet-b0, 53.04) (EfficientNet-b5, 54.28) (ShuffleNet-V2, 55.24)};
            \addplot [fill=pink] coordinates {(EfficientNet-b0, 53.38) (EfficientNet-b5, 54.86) (ShuffleNet-V2, 56.11)};
            \addplot [fill=cyan] coordinates {(EfficientNet-b0, 56.98) (EfficientNet-b5, 58.48) (ShuffleNet-V2, 60.45)};
            \end{axis}
        \end{tikzpicture}
    \end{minipage}
    \hfill
    \begin{minipage}{0.48\textwidth}
        \centering
        \begin{tikzpicture}
            \begin{axis}[
                ybar,
                bar width=0.2cm,
                width=0.9\textwidth,
                height=5cm,
                ylabel={Mean Precision($\%$)},
                ymin=40,
                symbolic x coords={EfficientNet-b0, EfficientNet-b5, ShuffleNet-V2},
                xtick=data,
                legend style={at={(0.5,-0.2)}, anchor=north, legend columns=-1},
                enlarge x limits=0.15,
            ]
            \addplot [fill=gray] coordinates {(EfficientNet-b0, 50.26) (EfficientNet-b5, 50.36) (ShuffleNet-V2, 52.68)};
            \addplot [fill=orange] coordinates {(EfficientNet-b0, 52.87) (EfficientNet-b5, 54.19) (ShuffleNet-V2, 54.87)};
            \addplot [fill=pink] coordinates {(EfficientNet-b0, 53.95) (EfficientNet-b5, 54.65) (ShuffleNet-V2, 56.54)};
            \addplot [fill=cyan] coordinates {(EfficientNet-b0, 57.01) (EfficientNet-b5, 58.39) (ShuffleNet-V2, 60.57)};
            \end{axis}
        \end{tikzpicture}
    \end{minipage}

    \vspace{0.2cm}

    \begin{minipage}{0.48\textwidth}
        \centering
        \begin{tikzpicture}
            \begin{axis}[
                ybar,
                bar width=0.2cm,
                width=0.9\textwidth,
                height=5cm,
                ylabel={Mean Recall($\%$)},
                ymin=40,
                symbolic x coords={EfficientNet-b0, EfficientNet-b5, ShuffleNet-V2},
                xtick=data,
                legend style={at={(0.5,-0.2)}, anchor=north, legend columns=-1},
                enlarge x limits=0.15,
            ]
            \addplot [fill=gray] coordinates {(EfficientNet-b0, 49.91) (EfficientNet-b5, 51.05) (ShuffleNet-V2, 53.64)};
            \addplot [fill=orange] coordinates {(EfficientNet-b0, 53.12) (EfficientNet-b5, 54.40) (ShuffleNet-V2, 55.08)};
            \addplot [fill=pink] coordinates {(EfficientNet-b0, 54.02) (EfficientNet-b5, 54.46) (ShuffleNet-V2, 55.98)};
            \addplot [fill=cyan] coordinates {(EfficientNet-b0, 56.87) (EfficientNet-b5, 58.41) (ShuffleNet-V2, 60.45)};
            \end{axis}
        \end{tikzpicture}
    \end{minipage}
    \hfill
    \begin{minipage}{0.48\textwidth}
        \centering
        \begin{tikzpicture}
            \begin{axis}[
                ybar,
                bar width=0.2cm,
                width=0.9\textwidth,
                height=5cm,
                ylabel={Mean F1 Score($\%$)},
                ymin=40,
                symbolic x coords={EfficientNet-b0, EfficientNet-b5, ShuffleNet-V2},
                xtick=data,
                legend style={at={(0.5,-0.2)}, anchor=north, legend columns=-1},
                enlarge x limits=0.15,
            ]
            \addplot [fill=gray] coordinates {(EfficientNet-b0, 48.92) (EfficientNet-b5, 50.03) (ShuffleNet-V2, 53.46)};
            \addplot [fill=orange] coordinates {(EfficientNet-b0, 53.26) (EfficientNet-b5, 54.31) (ShuffleNet-V2, 54.96)};
            \addplot [fill=pink] coordinates {(EfficientNet-b0, 53.86) (EfficientNet-b5, 54.55) (ShuffleNet-V2, 56.44)};
            \addplot [fill=cyan] coordinates {(EfficientNet-b0, 57.15) (EfficientNet-b5, 58.27) (ShuffleNet-V2, 60.47)};

            \end{axis}
        \end{tikzpicture}
    \end{minipage}
    
\vspace{0.2cm}

    \begin{minipage}{0.96\textwidth}
        \centering
        \begin{tikzpicture}
            \begin{axis}[
                width=\textwidth,
                height=2.0cm,
                legend style={at={(0.5,-0.2)}, anchor=north, legend columns=-1},
                legend cell align=center,
                area legend,
                bar width=0.2cm,
                ybar stacked,
                ylabel={},
                xmin=0,
                xmax=1,
                ymin=0,
                ymax=0.1,
                xtick=\empty,
                ytick=\empty,
                axis lines=none,
            ]
            
            \addlegendimage{fill= gray, draw=black}
            \addlegendentry{baseline}

            \addlegendimage{fill= orange, draw=black}
            \addlegendentry{baseline+CIBM}

            \addlegendimage{fill= pink, draw=black}
            \addlegendentry{baseline+CS}

            \addlegendimage{fill= cyan, draw=black}
            \addlegendentry{baseline+CIBM+CS}

            \end{axis}
        \end{tikzpicture}
    \end{minipage}

    \caption{Performance comparison of different classification networks as baseline combined with our proposed plug-and-play modules.}
    \label{fig:ablation comparison}
    
\end{figure*}
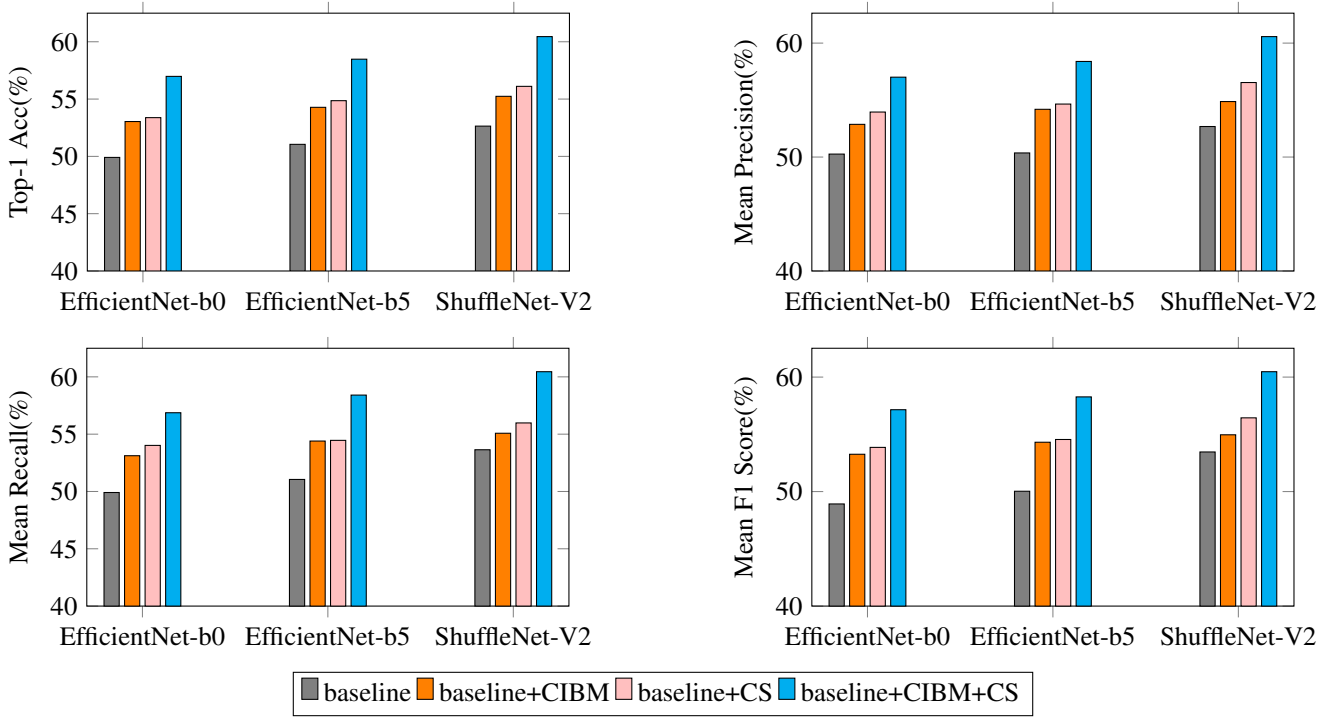

\subsection{\textbf{Effect of Super-Resolution Module}}
\label{sec:abla ddpm}
To illustrate the effectiveness of our proposed DDPM-based image super-resolution method for fine building classification, we trained the classification network with data generated by other competing super-resolution methods and compared the classification results while keeping the classification network unchanged. As shown in Table \ref{tab:abla ddpm}, all the methods to be compared were trained on the same dataset and migrated for the application to satellite images for fine building classification.  Results after processed by the four super-resolution methods show that our DDPM-based method outperforms the other three ones significantly after migration processing, although the method based on interpolation is more straightforward and efficient to implement. The experimental results show that it is difficult for low resolution overhead images to classify fine-grained buildings, and the DDPM-based method is practical and effective in terms of improving the image quality and information details contained.

\vspace*{-0.08cm}
\begin{table*}[h]
 \caption{Comparison of results from different methods, $\uparrow$ means higher is better.}
    \centering
    \small
\scalebox{.95}
    {    
    \begin{tabular}{l@{\hspace{.15cm}}c@{\hspace{.15cm}}c@{\hspace{.12cm}}c@{\hspace{.15cm}}c@{\hspace{.12cm}}c@{\hspace{.15cm}}c@{\hspace{.12cm}}c@{\hspace{.15cm}}c@{\hspace{.12cm}}c@{\hspace{.12cm}}c@{\hspace{.12cm}}c}
    \hline
    \text{Method} & \text{Top-1 Acc} $\uparrow$ & $\Delta$ & \text{Top-5 Acc} $\uparrow$ & $\Delta$ & \text{Mean Precision} $\uparrow$ & $\Delta$ &  \text{Mean Recall} $\uparrow$ & $\Delta$ & \text{Mean F1 Score} $\uparrow$ & $\Delta$  \\
    \hline
    \text{Eb0(EfficientNet-b0)} & 49.91 & 0 & 90.68 & 0 & 50.26 & 0 & 49.91 & 0 & 48.92 & 0 \\
    \text{Eb0+CIBM} & 53.04 & ${+6.3\%}$ & 91.09 & ${+0.4\%}$ &52.87 & ${+5.2\%}$ & 53.12 & ${+6.4\%}$ & 53.26 & ${+8.8\%}$ \\
    \text{Eb0+CS} & 53.38 & ${+6.9\%}$ & 90.87 & ${+0.2\%}$ &53.95 & ${+7.3\%}$ & 54.02 & ${+8.2\%}$ & 53.86 & ${+10.1\%}$ \\
    \textbf{Eb0+CIBM+CS} & \textbf{56.98} & ${+14.1\%}$ &\textbf{91.45} & ${+0.8\%}$ &\textbf{57.01} & ${+13.4\%}$  &\textbf{56.87} & ${+13.9\%}$ & \textbf{57.15} & ${+16.8\%}$\\
    \hline
    
    \text{Eb5(EfficientNet-b5)} & 51.05 & 0 & 90.91  & 0 & 50.36  & 0 & 51.05  & 0 & 50.03  & 0 \\
    \text{Eb5+CIBM} & 54.28 & ${+6.2\%}$ & 91.47 & ${+0.6\%}$ & 54.19 & ${+7.6\%}$ & 54.40 & ${+6.5\%}$ & 54.31 & ${+8.5\%}$ \\
    \text{Eb5+CS} & 54.86 & ${+7.4\%}$ & 91.00 & ${+0.01\%}$ & 54.65 & ${+8.5\%}$ & 54.46 & ${+6.6\%}$ & 54.55 & ${+9.0\%}$ \\
    \textbf{Eb5+CIBM+CS} & \textbf{58.48}& ${+14.5\%}$ & \textbf{92.75} & ${+2.0\%}$ & \textbf{58.39}& ${+15.9\%}$ & \textbf{58.41} & ${+14.4\%}$ & \textbf{58.27}& ${+16.4\%}$ \\
    \hline

    \text{Sv2(ShuffleNet-V2)} & 52.64  & 0 & 90.95  & 0 & 52.68  & 0 & 53.64  & 0 & 53.46 & 0 \\
    \text{Sv2+CIBM} & 55.24 & ${+4.9\%}$ & 91.12 & ${+0.2\%}$ &54.87 & ${+4.2\%}$ & 55.08 & ${+2.7\%}$ & 54.96 & ${+2.8\%}$ \\
    \text{Sv2+CS} & 56.11 & ${+6.6\%}$ & 92.08 & ${+1.2\%}$ & 56.54 & ${+7.3\%}$ & 55.98 & ${+4.3\%}$ & 56.44& ${+5.6\%}$ \\
    \textbf{Sv2+CIBM+CS} & \textbf{60.45} & ${+14.8\%}$ & \textbf{93.50} & ${+2.8\%}$ & \textbf{60.57} & ${+14.9\%}$ & \textbf{60.45}& ${+12.6\%}$ & \textbf{60.47} & ${+13.1\%}$ \\
    \hline
    \end{tabular}
    }
    \label{tab:abla phase2}
    \vspace*{0.1cm}
    \vspace*{-0.1cm}
\end{table*}

\subsection{\textbf{Effect of Building Classification Module}}
\label{sec:abla phase2}
To illustrate the effectiveness of our proposed contrastive supervision network and CIBM for building classification, we compared the results after incoperating these methods, as shown in Table \ref{tab:abla phase2} and Fig.\ref{fig:ablation comparison}.
In order to verify the effectiveness and plug-and-play ubiquity of our proposed CIBM and CS, ablation experiments are conducted using SOTA baseline networks, noting that all experiments were performed on the results of phase1 processing to ensure data consistency. In the experiments, CIBM and CS are added to baseline networks one by one for the performance test. The $\Delta$ right column of each metric denotes the percentage of relative increment. Specifically, the data in the table can be roughly summarized to show that respective effect of CIBM and CS on the performance of different networks is not identical, and there is a preference for certain metrics, but also some common features. For example, the combined use of CS and CIBM resulted in a better improvement in the network performance than the simple sum of improvements obtained by using them separately, suggesting a positive coupling between the proposed CS and CIBM. This is probably due to the fact that samples selected by CIBM during the training changed the internal feature distance of categories within ImageNet-1K. It is beneficial information for contrastive supervision, which in turn back-propagates to adjust the model to better fit the current training dataset, allowing the CS and CIBM to be more effective. This is a very interesting phenomenon, as at beginning of designing the network, we did not expect them to achieve a 1+1$>$2 result.

\section{Conclusion}
\label{sec:conlusion}
The fine classification of urban buildings based on remote sensing images is a popular research topic, as the results are useful to giving a good idea of the economic, industrial and even population distribution within a city. This is essential for urban planning, road construction, etc. However, there are two main challenges: 1. the low resolution of overhead views from high altitude remote sensing satellites, and 2. the strong variation in the number of building instances of different types, making the class imbalance a severe problem in the acquired training data. To address these two problems, we develop a two-phase strategy for fine-grained building classification from coarse overhead images. In the first phase, we design a model migration-based DDPM method to enhance the low-resolution satellite images, and in the second phase, we design a category information balanced module (CIBM) and contrastive supervision (CS) to improve the performance of the fine-grained building classification network. We achieved promising results on a Google Earth-based intercepted satellite image dataset, while full ablation experiments to verify the effectiveness of improvements were conducted. Our research contributes to the field of urban analysis by providing a practical and efficient solution for fine classification of urban buildings in large-scale challenging scenarios using satellite images. The proposed approach can serve as a valuable tool for urban planners, aiding in the understanding of economic, industrial, and population distribution within cities and regions, ultimately facilitating informed decision-making processes in urban development and infrastructure planning.

\section*{Acknowledgements}

This work was supported by National Natural Science Foundation of China (Grant No.42171361) and by a grant from the
Research Grants Council of the Hong Kong Special Administrative Region, China (Project No. PolyU 15215023). This work was also funded by the research project (Project Number: 2021.A6.184.21D) of the Public Policy Research Funding Scheme of The Government of the Hong Kong Special Administrative Region. This work was partially supported by The Hong Kong Polytechnic University under Projects 1-YXAQ and Q-CDAU. Thanks also go to KaYee Ng from The Hong Kong Polytechnic University for her assistance in the raw data collection and illustration process.





\bibliographystyle{elsarticle-harv} 
\bibliography{Re_building_classification, Re_super_resolution}





\end{document}